\pdfoutput=1
\documentclass[11pt]{article}

    \usepackage[preprint]{neurips2023}

\usepackage{times}
\usepackage{latexsym}

\usepackage[T1]{fontenc}
\usepackage[utf8]{inputenc}

\usepackage{microtype}

\usepackage{inconsolata}

\usepackage{graphicx} 
\usepackage{float} 
\usepackage{hyperref}
\usepackage{url}
\usepackage{booktabs}
\usepackage{multirow}
\usepackage{graphics}
\usepackage{amsmath}
\usepackage{multicol}
\usepackage{lipsum}
\usepackage{mwe}
\usepackage{xcolor}
\usepackage[ruled,vlined]{algorithm2e}
\usepackage{bbm}
\usepackage{upgreek}
\usepackage{svg}
\usepackage{colortbl}
\usepackage{subcaption}
\usepackage{wrapfig}

\usepackage{amsmath,amsfonts,bm}

\def\eqref#1{equation~\ref{#1}}
\def\1{\bm{1}}

\DeclareMathAlphabet{\mathsfit}{\encodingdefault}{\sfdefault}{m}{sl}
\SetMathAlphabet{\mathsfit}{bold}{\encodingdefault}{\sfdefault}{bx}{n}

\DeclareMathOperator*{\argmax}{arg\,max}

\colorlet{lgreen}{green!35}
\colorlet{lblue}{blue!20}
\colorlet{lred}{red!35}
\colorlet{lyellow}{yellow!45}

\newcommand{\aaron}[1]{{\color{blue} [{\bf Aaron}: #1]}}
\newcommand{\hanjie}[1]{{\color{magenta} [{\bf Hanjie}: #1]}}
\newcommand{\zhiyuan}[1]{{\color{orange} [{\bf Zhiyuan}: #1]}}
\newcommand{\xiang}[1]{{\color{cyan} [{\bf Xiang}: #1]}}

\newcommand{\method}{\textsc{KNIFE}}
\newcommand{\methodsp}{\textsc{KNIFE} }

\newcommand{\eg}{\textit{e.g., }}
\newcommand{\ie}{\textit{i.e., }}

\newcommand{\knifeemoji}{\raisebox{-3pt}{\includegraphics[width=1em]{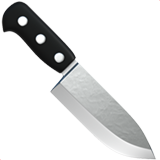}}}

\title{\knifeemoji \method: Distilling Reasoning \\ Knowledge From Free-Text Rationales}

\author{Aaron Chan$^{1}$\thanks{~~Equal contribution.} \hspace{3mm} Zhiyuan Zeng$^{2*}$ \hspace{3mm} Wyatt Lake$^{3}$ \\ \textbf{Brihi Joshi}$^{1}$ \hspace{3mm} \textbf{Hanjie Chen}$^{4}$ \hspace{3mm} \textbf{Xiang Ren}$^{1}$ \vspace{1mm} \\
$^{1}$University of Southern California \hspace{1mm} $^{2}$Tsinghua University
\\
$^{3}$Harvard-Westlake School \hspace{1mm} $^{4}$University of Virginia
\\
\small{\texttt{\{chanaaro, brihijos, xiangren\}@usc.edu}}$^{1}$
\hspace{3mm}
\small{\texttt{zengzy20@mails.tsinghua.edu.cn}}$^{2}$
\\
\small{\texttt{wlake2@hwemail.com}}$^{3}$
\hspace{3mm}
\small{\texttt{hc9mx@virginia.edu}}$^{4}$
}

\begin{document}
\maketitle

\begin{abstract}
Language models (LMs) have yielded impressive results on many language reasoning tasks, but their unexpected errors raise doubts about their reasoning abilities.
In light of this, there is growing interest in finetuning/prompting LMs with both task instances and their associated free-text rationales (FTRs), which explain the correct reasoning process for predicting the correct task output (\ie how to be ``right for the right reasons'').
However, existing finetuning methods fail to improve LM performance, while prompting needs prohibitively large (\ie >50B) LMs to work well.
% Whereas an FTR provides instance-level reasoning knowledge, a set of FTRs can collectively convey task-level reasoning knowledge that generalizes to unseen instances.
We propose \method, which shows that reasoning knowledge can be effectively distilled from FTRs into a small (\ie <1B) LM and improve the LM's performance.
First, \methodsp finetunes a teacher LM (given task input and FTR) to predict the task output, transferring reasoning knowledge from the FTRs to the teacher's hidden states.
Second, \methodsp finetunes a student LM (given task input only) such that its hidden states are aligned with the teacher's.
Thus, the student is endowed with reasoning knowledge but can be used for inference without direct FTR input.
On two question-answering datasets, \methodsp outperforms various finetuning and prompting baselines in fully-supervised and low-resource settings.
Also, we observe that FTR quality is crucial to \method's performance.
\end{abstract}

\renewcommand{\aaron}[1]{}
\renewcommand{\xiang}[1]{}
\renewcommand{\hanjie}[1]{}
\renewcommand{\zhiyuan}[1]{}

\section{Introduction} 
\label{sec:intro}

Whereas conventional supervised learning only gives feedback on a language model's (LM's) task output correctness, \textit{explanation tuning} aims to teach LMs to be ``right for the right reasons'' \citep{ross2017right} by also providing them with explanations of the correct reasoning process behind a given correct output~\citep{narang2020wt5, hase2021can, joshi2022ertest}.
In particular, there is growing interest in learning from \textit{free-text rationales} (FTRs), which use natural language to verbally explain the correct reasoning process for solving a given task instance~\citep{camburu2018snli, rajani2019explain, narang2020wt5, wei2022chain}.

\begin{wrapfigure}{R}{0.5\textwidth}
    \centering
    \centering
    \vspace{-0.5cm}
    \includegraphics[width=0.5\columnwidth]{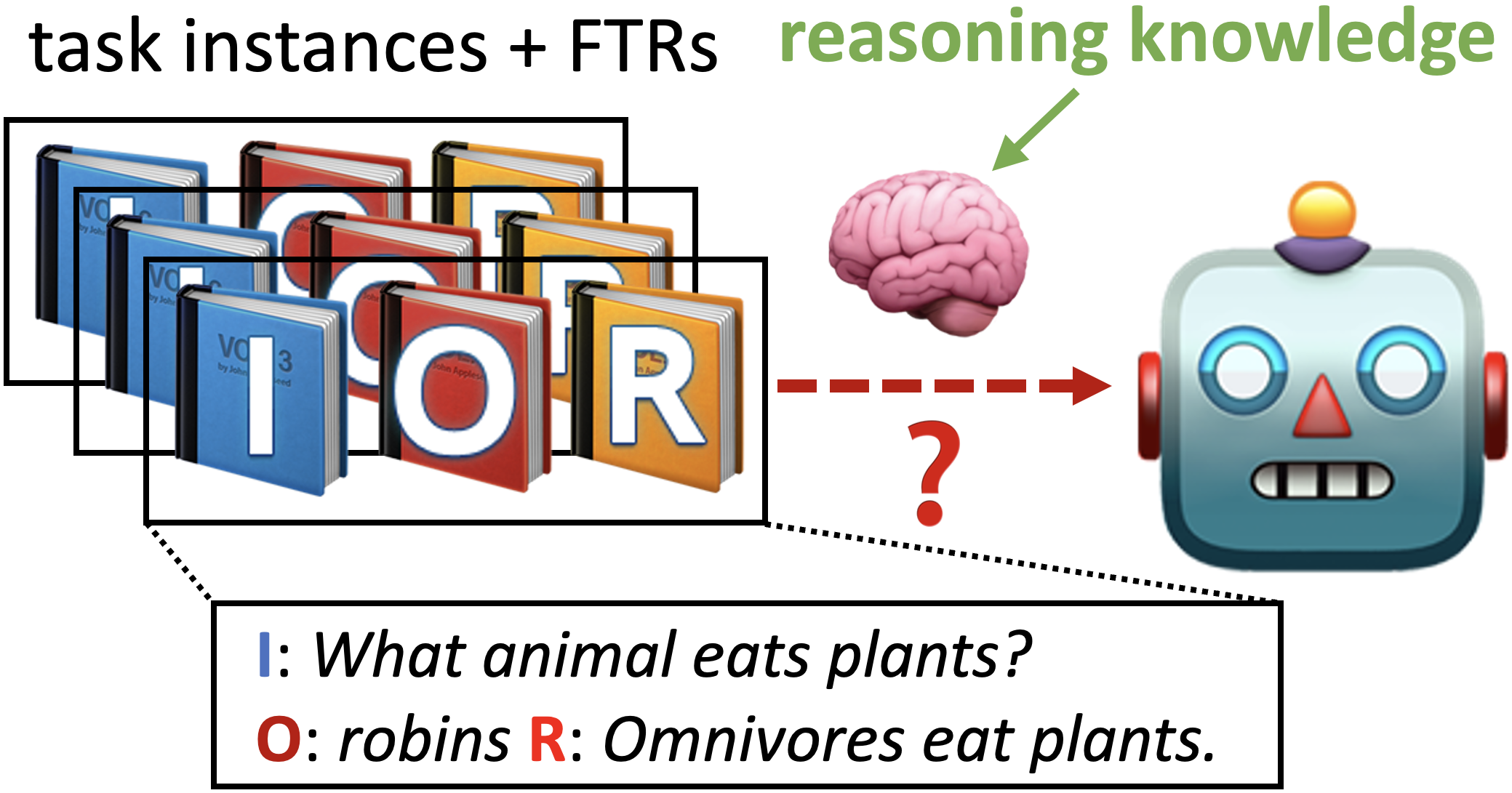}
    % \caption {\textbf{FTRs and meta-reasoning knowledge.} Meta-reasoning knowledge is a general but latent reasoning way to solve the considered task. Each FTR is an instantiation of the meta-reasoning knowledge for the corresponding instance. Conversely, a set of FTRs collectively conveys meta-reasoning knowledge through their general reasoning patterns. We aim to extract the meta-reasoning knowledge from the FTRs of training instances and inject it into an LM.}
    \caption{\textbf{Reasoning Knowledge From FTRs.} A free-text rationale (FTR) explains the correct reasoning process for solving a given task instance. Meanwhile, a set of FTR-augmented task instances (\ie \textbf{I}nput, \textbf{O}utput, FT\textbf{R}) can collectively provide latent reasoning knowledge for solving the task in general. Nonetheless, it remains unclear how to effectively inject this knowledge into LMs to improve their generalization performance.}
    \vspace{-0.4cm}
    \label{fig:ftr-knowledge-transfer}
\end{wrapfigure}

Among explanation tuning methods, the self-rationalization paradigm has been most successful in improving LM task performance~\citep{hase2021can,wei2022chain,lampinen2022can}.
Here, the LM is prompted or finetuned to jointly generate the task output and FTR~\citep{liu2018towards, narang2020wt5, marasovic2021few, brahman2021learning, wei2022chain, zelikman2022star}.
Although prompted self-rationalization can improve task performance in certain situations \citep{wei2022chain, lampinen2022can, zelikman2022star}, prompting typically requires prohibitively large-scale (\ie >50B) LMs to work well ~\citep{wei2022emergent}.
Meanwhile, small-scale (\ie <1B) LMs are suitable for finetuning, but finetuned self-rationalization is mostly used in the context of explainability and fails to consistently improve task performance~\citep{hase2021can, zhang2023multi}.
Since few AI researchers or practitioners have the computational resources to freely experiment with their own large-scale LMs \citep{zhao2023survey}, how can we use FTRs to improve the task performance of small-scale LMs?

Since an individual FTR explains the reasoning process for solving a single task instance, it only provides \textit{instance-level} reasoning knowledge.
Thus, given a set of task instances that sufficiently characterizes the task, it follows that a set of FTRs for these instances can collectively capture \textit{task-level} reasoning knowledge that generalizes to unseen task instances.
% We denote this implicit task-level reasoning knowledge as \textbf{\textit{meta-reasoning knowledge}}.
With this in mind, our goal is to finetune a small-scale LM by extracting reasoning knowledge from a set of FTRs, then injecting this knowledge into the LM to guide its reasoning and improve its inference performance (Fig.~\ref{fig:ftr-knowledge-transfer}).
We assume that all FTRs convey a correct reasoning process and are inaccessible to the LM during inference, since correct FTRs necessarily entail knowledge of the correct task output.

%The hope is that the LM will be able to leverage the meta-reasoning knowledge to solve unseen instances at inference time.

%While an individual FTR provides an \textbf{instance-level} reasoning process for a specific instance, our intuition is that the FTR of this instance is an \textbf{instantiation} of the \textbf{task-level} \textit{meta-reasoning knowledge}, which is a general but latent reasoning way to solve the considered task.
%From this perspective, a set of FTRs \textbf{collectively conveys} task-level meta-reasoning knowledge through their general reasoning patterns. 
%We assume having access to FTRs of training instances, while unrealistic is access to FTRs of inference instances as annotation of gold FTRs entails annotation of correct output.
%With this in mind, our goal is to extract meta-reasoning knowledge from a set of its instantiations (\ie FTRs) for training instances and effectively inject this knowledge into an LM, in order to improve its task performance (Fig.~\ref{fig:ftr-knowledge-transfer}). The LM can then solve unseen instances during inference time guided by the extracted meta-reasoning knowledge.

% \aaron{Finish writing below}

We propose \textbf{KN}owledge D\textbf{I}stillation From \textbf{F}ree-Text Rational\textbf{E}s (\knifeemoji \textbf{\method}), which shows that reasoning knowledge can be effectively distilled from FTRs into a small LM and improve the LM's performance.
Unlike prior works, \methodsp compresses FTR reasoning knowledge from its natural language form into a specific set of LM hidden states, making it easier to control how this knowledge is transferred.
First, \methodsp finetunes a teacher LM to predict the task output, given both the task input and an FTR.
This aggregates reasoning knowledge across all finetuning instances' FTRs, then transfers it to the teacher's hidden states.
Second, \methodsp finetunes a student LM, given only the task input, so that its hidden states are aligned with the teacher's.
By treating the teacher's hidden states as soft labels, \methodsp distills reasoning knowledge via such soft labels from the teacher to the student, which can then be used for inference without FTR input or FTR generation.

Across two QA datasets (OpenBookQA, StrategyQA) and two small-scale LM architectures (T5-Base, T5-Large), we show that \methodsp outperforms various finetuning and prompting baselines on both fully-supervised and low-resource settings, using either human-annotated or model-generated FTRs (\textsection \ref{sec:exp:main}).
% Also, \methodsp can outperform baselines in low-resource settings (\ie few training instances), further verifying \method's ability to improve LM generalization (\textsection \ref{sec:exp:low}) \zhiyuan{don't need that}.
% we show that \methodsp significantly outperforms existing FTR-based EBL methods, in both fully-supervised and low-resource settings.
% Also, \methodsp can outperform baselines on out-of-distribution (OOD) test sets, further verifying \method's ability to improve LM generalization. \aaron{Check this claim} \zhiyuan{Don't think we improve OOD performance. We improve performance in the low-resource setting.}
Furthermore, we validate~\methodsp design choices via extensive ablation studies (\textsection \ref{sec:exp:abl}).
% Moreover, we analyze the relationship between student LM performance and FTR quality, identifying FTR properties that are most helpful for \method. \zhiyuan{For the ACL submission, we should justify it is reasonable that our method is influenced by rationale quality. Otherwise we may be challenged by reviewers who say our method is not universal enough.} \aaron{Add more findings}
% Moreover, we analyze the impact of FTR quality on \method's distillation performance (\textsection \ref{}). \zhiyuan{case study sec?}
Finally, we analyze~\method's failure modes on two additional datasets (ECQA, QuaRTz) and identify FTR quality as a critical factor in \method's performance (\textsection \ref{sec:exp:failure}).

\section{Related Work}
\label{sec:rw}

% \aaron{Compress this}

% \paragraph{Explanation-Based Learning}
\vspace{-0.1cm}
\paragraph{Explanation Tuning}
There are three main existing paradigms for FTR-based explanation tuning: \textit{self-rationalization}, \textit{input augmentation}, and \textit{pipeline rationalization} (Fig. \ref{fig:ftr-paradigms}).
In self-rationalization, the LM is finetuned or prompted to generate both the task output and FTR~\citep{liu2018towards, narang2020wt5, brahman2021learning, marasovic2021few, li2022explanations, wei2022chain, zelikman2022star, lampinen2022can, majumder2022knowledge}.
However, finetuning struggles to improve LM performance~\citep{hase2021can} and prior works on it focus on explainability or FTR generation capability.
Prompting requires large and often prohibitively large LMs to work well.

Besides self-rationalization, the other two paradigms struggle to improve task performance due to some intrinsic issues.
% In input augmentation, the LM is finetuned to generate the task output given both the task input and an FTR \citep{sun2022investigating, wang2022pinto, wiegreffe2021measuring, hase2020leakage}.
% \citep{sun2022investigating, wang2022pinto, wiegreffe2021measuring, hase2020leakage}.
% Still, this either assumes access to gold (or large-LM-generated) FTRs during inference, or introduces an input distribution shift between training and inference when FTRs are unavailable during inference.
Methods under \textit{input augmentation} paradigm use FTRs as additional inputs~\citep{rajani2019explain, hase2020leakage, kumar2020nile, wiegreffe2021measuring}.
They need to resolve the input distribution shift issue, which occurs when incorporating FTRs into inputs during training but not during inference, and thus struggle to improve task performance.
% However, for self-rationalization, finetuning may create conflict between the task and FTR objectives, while prompting requires very large LMs to work well.
% \zhiyuan{Do you think we can use "objective" here? "loss" may sound too technical.}
% (esp. if the generated FTR does not support the LM's predicted task output \zhiyuan{Not sure whether this esp is necessary.})
Some other works explored the \textit{pipeline rationalization} paradigm, where a finetuned rationalizing LM first generates the FTR and then a finetuned reasoning LM predicts the task output given the generated FTR~\citep{rajani2019explain, kumar2020nile, hase2020leakage, wiegreffe2021measuring}.
However, the generated FTR could be poor, omitting critical information from the task input or introducing irrelevant and even incorrect information, which can hurt task performance~\citep{huang2022large, magister2022teaching, li2022explanations}.
Besides, the generated FTR forms a non-differentiable path, which complicates end-to-end training.

\paragraph{Knowledge Distillation}
% Knowledge distillation has been widely used for developing efficient deep neural networks by transferring the knowledge from a larger teacher model to a smaller student model \citep{hinton2015distilling, sanh2019distilbert, jiao2020tinybert}. 
Knowledge distillation has been widely used to transfer knowledge from a larger teacher to a smaller student model~\citep[etc]{hinton2015distilling, sanh2019distilbert, jiao2020tinybert, mirzadeh2020improved}. 
% Previous works have explored knowledge distillation from different perspectives, such as teacher-student architectures \citep{mirzadeh2020improved}, distillation algorithms \citep{passalis2018learning}, knowledge types \citep{cheng2020explaining}, data efficiency \citep{ji2020knowledge}, etc. 
% The types of knowledge for distillation include the logits of teacher model \citep{hinton2015distilling, kim2018paraphrasing} or the representations extracted from the intermediate layers of teacher model \citep{romero2014fitnets, heo2019knowledge, tang2019distilling, sun2019patient}. 
% We follow another line of work that incorporate knowledge distillation with privileged information \citep{lopez2015unifying, vapnik2015learning, fukuda2017efficient, wang2018kdgan}.
Instead of aiming for this typical goal, \method~distills the FTR knowledge through the hidden states of a teacher model to a student model, which has no direct access to FTRs. Similar to the line of work that incorporates knowledge distillation with privileged information \citep{lopez2015unifying, vapnik2015learning, fukuda2017efficient, wang2018kdgan}, where student models benefit from privilege information, the student model in \method~essentially gains additional knowledge from FTRs rather than relying on the larger teacher model capacity. \citet{snell2022learning} propose to internalize the in-context learning ability such that the performance gains can keep without context tokens. It does not directly distill the knowledge from FTRs and requires dedicated prompt designs. \citet{shridhar2022distilling, magister2022teaching, namgyu2022reasoingteacher} propose to distill reasoning abilities from larger language models to smaller models. They require large-scale language models with such abilities, while \method~can work well with small models.
\section{Background} 
\label{sec:bg}

\paragraph{Problem Definition}
Given a task input $\mathbf{x}$ and a set of class labels $Y = \{\mathbf{y}_i\}$, the model's goal is to predict a score $\rho(\mathbf{x}, \mathbf{y}_i)$ for each $(\mathbf{x}, \mathbf{y}_i)$ pair, so that the predicted label $\hat{\mathbf{y}} = \argmax_{\mathbf{y}_i \in Y} \hspace{0.5mm} \rho(\mathbf{x}, \mathbf{y}_i)$ matches the gold (\ie correct) label $\mathbf{y}^* \in Y$.
We consider both multi-choice ($Y$ varies across task instances) and closed-set ($Y$ is fixed for all task instances) text classification.
% \zhiyuan{closed-label and open-label are two wired terms to me. I think people often say something like "text classification" and "multiple choice"? We can check more about that.}

% We refer to $y^*$ as the gold label for $x$.
% Let $x$ be the task input (\eg question).
% Let $y^* \in Y$ be the gold label (\eg correct answer), where $Y = \{y_i\}$ is the label space.

\paragraph{Free-Text Rationales}
A \textit{free-text rationale} (FTR) $\mathbf{r}$ for a pair $(\mathbf{x}, \mathbf{y}_i)$ is a natural language text that explains the reasoning process for predicting label $\mathbf{y}_i$ \citep{camburu2018snli, rajani2019explain}.
% Compared to extractive rationales \citep{sundararajan2017axiomatic, li2016understanding, chan2022unirex}
FTRs could be intuitive to humans, reference things beyond the task input, and support high flexibility in content, style, and length \citep{wiegreffe2021measuring, chan2022frame}.
Recent works have explored generating FTRs to explain LM behavior \citep{camburu2018snli, narang2020wt5, wei2022chain} and utilizing FTRs to improve LMs \citep{sun2022investigating, wang2022pinto, li2022explanations}.
We assume each training instance $\mathbf{x}$ is accompanied by an annotated FTR $\mathbf{r}$ for $\mathbf{y}^*$, while impossible is access to FTRs for inference instances as FTR explains the reasoning process and thus indicates the gold label. In this setting, we aim to improve $\mathcal{F}$'s performance by these annotated FTRs.
\section{\method} 
\label{sec:method}

\begin{figure*}
    \centering
    \vspace{-0.8cm}
    \includegraphics[width=\textwidth]{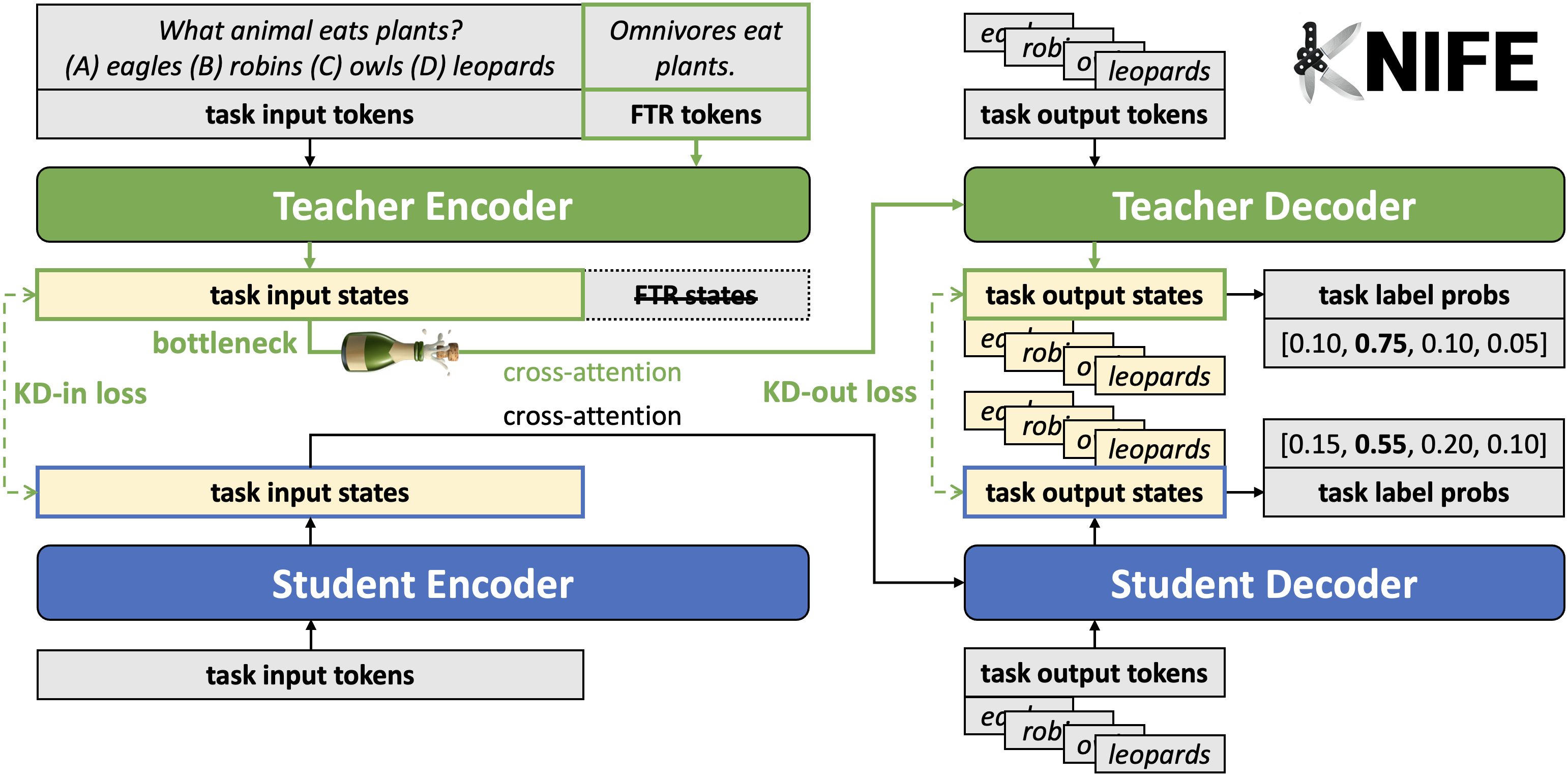}
    \caption{\textbf{KNIFE Framework.}
    \methodsp distills reasoning knowledge from an FTR-augmented teacher LM (given task input and FTR) to a student LM (given task input) that is used for inference. The teacher has a \textit{bottleneck}, which masks out all FTR states during cross-attention. As a result, the teacher's decoder must use only its task input (hidden) states to compute its task output (hidden) states, thus routing FTR knowledge to the task input/output states. Finally, FTR knowledge is distilled to the student by training the student so its task input/output states to align with the teacher's.}
    % \vspace{-0.6cm}
    \label{fig:knife}
\end{figure*}

% \hanjie{Would it be better to introduce the KD (\method) framework first, and then introduce each part of LM designs? I was kind of overwhelmed by the design details before I got a full picture of the \method~framework.} \zhiyuan{@Hanjie do you think the current modification is OK?}

% \hanjie{I feel the following paragraph has been repeated multiple times so far. We may omit it here?}
% During task finetuning, existing works have attempted to learn from FTRs by appending FTRs to the LM's input \citep{sun2022investigating, wang2022pinto, wiegreffe2021measuring, hase2020leakage} or jointly training the LM to generate FTRs \citep{narang2020wt5, rajani2019explain, brahman2021learning, li2022explanations, hase2020leakage}.
% Many existing works aim to learn from FTRs by simply appending them to the LM's input or target output during training. \aaron{Add citations}
% Alternatively, a few prior works use KD to learn from FTRs \citep{snell2022learning, shridhar2022distilling}.
% However, these works suffer from input distribution shift or conflicting learning objectives.

% Existing works aim to learn from FTRs by using approaches like input augmentation, self-rationalization, or pipeline rationalization.
% However, such approaches may hurt task performance or require prohibitively large LMs to work well.
We propose~\method, an approach to extract reasoning knowledge from FTRs of training instances and inject it into an LM (Fig. \ref{fig:knife}) by knowledge distillation (KD).
First, \methodsp finetunes a teacher LM to predict the task output, taking as input the concatenation of task input and the FTR.
Then, \methodsp finetunes a student LM given only the task input, so that its task encoder/decoder hidden states are aligned with the teacher's.
As the teacher and student have different structures of encoder hidden states, the teacher LM has a \textit{bottleneck} design, where the encoder hidden states upon the FTR tokens are masked out in the cross-attention.
We are going to elaborate on the method design.

\subsection{LM Designs}
\label{sec:method:lm}

\methodsp consists of a student LM and a teacher LM.
They share the same encoder-decoder Transformer architecture and the basic design, with differences in the input and cross-attention mechanism.
We are going to first present the basic design, followed by the student and teacher LM designs respectively.
% \aaron{Figure out better names for: (1) task input states and (2) task output states}

\subsubsection{Basic Design}

% \hanjie{What do you mean by ``Basic Design"? It looks like ``Preliminaries" or may be included into ``Background"?}
% Following prior works,
We use the encoder-decoder Transformer \citep{vaswani2017attention} architecture for both teacher and student LMs \citep{raffel2020exploring, narang2020wt5}.
Building upon \textsection \ref{sec:bg}, we feed the input to the LM's encoder and separately feed each label $\mathbf{y}_i$ to the decoder. 
For each label $\mathbf{y}_i$, we get the conditional probabilities of all its tokens and then take the average log probability as its score $\rho(\mathbf{x}, \mathbf{y}_i)$.
% The scores are also logits for the cross-entropy loss calculation in training.
This practice is also adopted by \citet{shwartz2020unsupervised} and \citet{wang2022pinto}.
Formally speaking, each label is denoted as $n_{y_i}$-token sequence $\mathbf{y}_i = [y_i^{(1)}, y_i^{(2)}, \dots , y_i^{(n_{y_i})}] \in Y$. By separately teacher-forcing each $\mathbf{y}_i$ to the decoder, we get a conditional probability $P(y_i^{(j)} \hspace{0.5mm} | \hspace{0.5mm} y_i^{(1)}, \dots , y_i^{(j-1)}, \mathbf{x})$ for each token $y_i^{(j)}$ in $\mathbf{y}_i$.
% To avoid this issue, we instead produce a dedicated decoder output for each label sequence $\mathbf{y}_i$, by separately teacher-forcing each $\mathbf{y}_i$ to be $\mathcal{F}_{\text{dec}}$'s input  \citep{wang2022pinto}.
% This gives us a conditional probability $P(y_i^{(j)} \hspace{0.5mm} | \hspace{0.5mm} y_i^{(1)}, ... , y_i^{(j-1)}, \mathbf{x})$ for each token $y_i^{(j)}$ in $\mathbf{y}_i$. 
% Then, following \citet{wang2022pinto} and \citet{shwartz2020unsupervised}
We compute $\rho_i = \rho(\mathbf{x}, \mathbf{y}_i)$ as the score for $\mathbf{y}_i$, by aggregating these token probabilities as:
\begin{equation*}
   \rho_i=\frac{1}{n_{y_i}}\sum_{j=1}^{n_{y_i}} \log P(y_i^{(j)} \hspace{0.5mm} | \hspace{0.5mm} y_i^{(1)}, \dots , y_i^{(j-1)}, \mathbf{x}).
\end{equation*}
The predicted probability is calculated by the softmax function, \ie $P(\mathbf{y}_i \hspace{0.5mm} | \hspace{0.5mm} \mathbf{x})=e^{\rho_i}/\sum_{j=1}^{|Y|} \hspace{0.5mm} e^{\rho_j}$.
% Finally, we use the softmax function to normalize $\rho_i$ as label confidence probability $P(\mathbf{y}_i \hspace{0.5mm} | \hspace{0.5mm} \mathbf{x})=e^{\rho_i}/\sum_{j=1}^{|Y|} \hspace{0.5mm} e^{\rho_j}$.
% Given gold label $\mathbf{y}^*$, the goal of the downstream classification task is to train $\mathcal{F}$ such that $P(\mathbf{y}^* \hspace{0.5mm} | \hspace{0.5mm} \mathbf{x})$ is maximized.

\subsubsection{Student LM Design}
The input to the student is always the raw task input, and the student is used for inference.
Instead of training the student LM by the cross-entropy loss, we train it to align its encoder/decoder hidden states with those of the teacher LM (\textsection \ref{sec:method:kd}), which is trained before the training of the student.
We only consider the hidden states at the top layer in this work.
As the token prediction logits are calculated by the LM head taking decoder hidden states as input, we adopt the teacher's LM head for the student.
% After finetuning $\mathcal{T}$, \methodsp finetunes a student LM based on KD objectives, given only $\mathbf{x}$.
% Let $\mathcal{S}$ be the student LM, with encoder $\mathcal{S}_{\text{enc}}$ and decoder $\mathcal{S}_{\text{dec}}$.
% Since $\mathcal{S}$ does not have access to $\mathbf{r}$, $\mathcal{S}_{\text{enc}}$ only outputs task input states $\mathbf{e}_{\text{S}} = \mathcal{S}_{\text{enc}}(\mathbf{x}) = [\mathbf{e}_{\text{S}x}^{(1)}, \mathbf{e}_{\text{S}x}^{(2)}, ..., \mathbf{e}_{\text{S}x}^{(n_{x})}]$ and can be used for inference.
% Hence, $\mathcal{S}$ does not have a bottleneck.
% For each label $\mathbf{y}_i$, $\mathcal{S}_{\text{dec}}$ produces task output states $\mathbf{d}_{\text{S}y_i} = \mathcal{S}_{\text{dec}}(\mathbf{y}_i, \mathbf{e}_{S}) = [\mathbf{d}_{\text{S}y_i}^{(1)}, \mathbf{d}_{\text{S}y_i}^{(2)}, ..., \mathbf{d}_{\text{S}y_i}^{(n_{y_i})}]$.
% Unlike $\mathcal{T}$, since $\mathcal{S}$ does not have access to any FTR $\mathbf{r}$, so $\mathcal{S}$ follows the same basic LM design (\ie no FTR bottleneck) described earlier.

\subsubsection{Teacher LM Design}

\paragraph{Main Design}
We aim to first obtain a teacher LM that does reasoning following the FTRs so that its reasoning is guided by reasoning knowledge.
To achieve this, we feed the FTRs to the teacher LM. When training the teacher on training instances by task loss $\mathcal{L}_{\text{task}}=-\log P(\mathbf{y}^* \hspace{0.5mm} | \hspace{0.5mm} \mathbf{x})$, we take as its input the concatenation of the task input and the corresponding FTR.
Intuitively, feeding the FTR along with the task input to the teacher LM forces the teacher to reason in a way following the FTR.

Even though the teacher LM's reasoning on training instances is guided by FTRs and thus the reasoning knowledge, we are unable to use it for inference as unavailable are the FTRs of inference instances.
However, considering the calculation of its hidden states on a training instance is guided by the corresponding FTR, the hidden states store the FTR knowledge.
Following the intuition that a set of FTRs collectively conveys reasoning knowledge, the set of hidden states also collectively conveys reasoning knowledge.
Inspired by the works on knowledge distillation~\citep[etc]{hinton2015distilling, sanh2019distilbert, jiao2020tinybert}, the hidden states of the teacher are also soft labels, on which the student could be trained.
By aligning the student's hidden states on training instances with those of the teacher, the knowledge across all FTRs or hidden states is synthesized into the reasoning knowledge conveyed collectively by them, which is finally distilled into the student LM.

\paragraph{FTR Bottleneck}
In the Transformer architecture, there is a one-to-one correspondence between each token fed to the encoder/decoder and each encoder/decoder hidden state.
Each token corresponds to the state upon it.
For a specific training instance, the token sequence fed to the student's decoder is identical to that fed to the teacher's decoder.
Thus, the student and the teacher share the same structure of the decoder hidden states.
We train the student to align each decoder hidden state with the teacher's one upon the same token.

However, the token sequence fed to the student' encoder (the raw task input) is a proper prefix of that fed to the teacher's encoder (the task input concatenated with the FTR).
We call the hidden states upon the task input tokens \textit{task input states} and call those upon the FTR tokens \textit{FTR states}.
The student and the teacher share the same structure of the task input states but the student doesn't have FTR states as the teacher does.
If we just directly ignored the teacher's FTR states in KD, the information stored in the FTR states would be lost.
If we fed only raw task input to the teacher in KD, the teacher itself would not work due to the input distribution shift because the FTR is always appended to the input during the training of the teacher.

% \hanjie{Reviewers might have difficulty in understanding the ``input distribution shift".} \zhiyuan{@Hanjie Do you think "as ..." solves that?}

To address this, the teacher LM has a \textit{bottleneck} design in the cross-attention connecting its decoder to the encoder.
Here, the teacher's FTR states are masked out in cross-attention so that the decoder only has access to the task input states.
In this way, the training of the teacher funnels the FTR knowledge stored in the FTR token sequence to the task input states by self-attention in the encoder.

As cross-attention is the only path introducing information from the encoder to the decoder, the teacher's FTR states are completely useless now owing to the bottleneck design.
Therefore, we can ignore the teacher's FTR states in knowledge distillation as if there had only existed task input states.

\subsection{Knowledge Distillation}
\label{sec:method:kd}

As both encoder hidden states (task input states) and decoder hidden states store the reasoning knowledge, KD is done by training the student LM so that its encoder and/or decoder hidden states are aligned with the teacher's.
We now formally define the learning objectives.
Supposing $\mathbf{x}$ is $n_{x}$-token sequence, the teacher's and student's task input states are denoted as $[\mathbf{e}_{\text{T}x}^{(1)}, \mathbf{e}_{\text{T}x}^{(2)}, \dots, \mathbf{e}_{\text{T}x}^{(n_{x})}]$ and $[\mathbf{e}_{\text{S}x}^{(1)}, \mathbf{e}_{\text{S}x}^{(2)}, \dots, \mathbf{e}_{\text{S}x}^{(n_{x})}]$ respectively.
Let dist denote the mean squared error (MSE), a distance function.
Let $\mathcal{L}_{\text{KD-In}}$ denote \method's encoder hidden states based KD loss, which pushes the student's task input states ($\mathbf{e}_{\text{S}x}^{(j)}$) to be closer to the teacher's ($\mathbf{e}_{\text{T}x}^{(j)}$):
\begin{equation*}
    \mathcal{L}_{\text{KD-In}} = \frac{1}{n_x} \sum_{j=1}^{n_x} \text{dist}(\mathbf{e}_{\text{S}x}^{(j)}, \mathbf{e}_{\text{T}x}^{(j)})
\end{equation*}

\noindent
Similarly, for each label $\mathbf{y}_i$ (fed to the decoder separately), the teacher's and student's decoder hidden states are denoted as $[\mathbf{d}_{\text{T}y_i}^{(1)}, \mathbf{d}_{\text{T}y_i}^{(2)}, \dots, \mathbf{d}_{\text{T}y_i}^{(n_{y_i})}]$ and $[\mathbf{d}_{\text{S}y_i}^{(1)}, \mathbf{d}_{\text{S}y_i}^{(2)}, \dots, \mathbf{d}_{\text{S}y_i}^{(n_{y_i})}]$ respectively.
Let $\mathcal{L}_{\text{KD-Out}}$ denote \method's decoder hidden states (task output states) based KD loss, which pushes the student's decoder hidden states ($\mathbf{d}_{\text{S}y_i}^{(j)}$) to be closer to the teacher's ($\mathbf{d}_{\text{T}y_i}^{(j)}$):
\begin{equation*}
    \mathcal{L}_{\text{KD-Out}} = \frac{1}{\sum_{y_i \in Y}n_{y_i}} \sum_{y_i \in Y} \sum_{j=1}^{n_{y_i}} \text{dist}(\mathbf{d}_{\text{S}y_i}^{(j)}, \mathbf{d}_{\text{T}y_i}^{(j)})
\end{equation*}

\noindent
Finally, let $\mathcal{L}$ denote the total loss defined as $\mathcal{L} = \lambda_{\text{KD-In}} \mathcal{L}_{\text{KD-In}} + \lambda_{\text{KD-Out}} \mathcal{L}_{\text{KD-Out}}$, with loss weights $\lambda_{\text{KD-In}}\in\{0,1\}$ and $\lambda_{\text{KD-Out}}\in\{0,1\}$.
% with loss weights $\lambda_{\text{task}}$, $\lambda_{\text{KD-in}}$, and $\lambda_{\text{KD-out}}$, such that $\lambda_{\text{task}} + \lambda_{\text{KD-in}} + \lambda_{\text{KD-out}} > 0$.
% Note that the teacher LM is trained only with $\mathcal{L}_{\text{task}}$, whereas the student LM may or may not use $\mathcal{L}_{\text{task}}$.
% Note that the teacher LM is trained only with $\mathcal{L}_{\text{task}}$ (\ie $\lambda_{\text{task}} > 0$ and $\lambda_{\text{KD-in}} = \lambda_{\text{KD-out}} = 0$), whereas the student LM may or may not use $\mathcal{L}_{\text{task}}$ (\ie $\lambda_{\text{task}}, \lambda_{\text{KD-in}}, \lambda_{\text{KD-out}} \geq 0$).
% \hanjie{In Table 1, I do not see the results of KNIFE Student Finetuning (Task+In+Out)?}

% \subsection{Advantages}
% \methodsp has several advantages over prior works.
% \aaron{Describe advantages}
\section{Experiments} 
\label{sec:exp}

This section presents experiments.\footnote{We report the mean and standard deviation (std) accuracy over three random seeds for all results, using the format \texttt{mean} $\pm$ \texttt{std}.
For each table, we use horizontal lines to partition the table into various sub-tables.
Each sub-table contains results for methods that have comparable settings (\eg architecture).
That is, \texttt{Result1} should only be compared to \texttt{Result2} if there is no horizontal line separating them in the table.
In each sub-table, we highlight the best-performing method in \colorbox{lred}{red} and the second-best performing method in \colorbox{lblue}{blue}.}
First, in both fully-supervised and low-resource settings, \methodsp outperforms various baselines, using either human-annotated FTRs or model-generated FTRs (\textsection \ref{sec:exp:main}).
% Second, we show that \methodsp can also outperform baselines in low-resource settings (\ie few training instances), further verifying \method's ability to improve LM generalization (\textsection \ref{sec:exp:low}).
Second, we validate our \methodsp design choices via extensive ablation studies (\textsection \ref{sec:exp:abl}).
Third, we analyze \method's failure modes and identify FTR quality as critical to \methodsp performance (\textsection \ref{sec:exp:failure}).
% ... (\textsection \ref{sec:exp:failure}) \aaron{Update this} 

% \subsection{Evaluation Protocol}

\subsection{Datasets}
We consider both multi-choice and closed-set text classification.
% Since FTRs are commonly annotated for QA datasets \citep{wiegreffe2021teach}, we focus on three QA datasets: OBQA, StrategyQA, and QuaRTz.
% Since FTRs are commonly annotated for question-answering (QA) datasets \citep{wiegreffe2021teach}, 
We focus on two question-answering (QA) datasets: OpenBookQA (OBQA) and StrategyQA.\footnote{Since StrategyQA does not provide public test set labels, we use the data split from~\citet{wang2022pinto}.}
% \zhiyuan{remember to mention StrategyQA in-house split}
OBQA~\citep{mihaylov2018can} is a 4-choice QA dataset that simulates science exams.
% In OBQA, each question requires both scientific knowledge and commonsense reasoning to solve.
StrategyQA~\citep{geva2021did} is a boolean QA dataset that requires multi-hop reasoning.
% In StrategyQA, each question must be solved by inferring the required reasoning steps implicit in the question.
% QuaRTz \citep{tafjord2019quartz} is a multi-choice (\ie two-choice) QA dataset that tests the ability to reason about qualitative relationships.
% In QuaRTz, each question involves open-domain relationships and requires general qualitative knowledge to solve. 
% For all datasets, performance is measured using accuracy.

% \subsubsection{Presentation of Results}
% \zhiyuan{How about move Results Reporting to the table's caption? It is a little waste of space to illustrate that in the main text?}
% For all results, we report the mean and standard deviation (std) accuracy over three random seeds, using the format \texttt{mean} $\pm$ \texttt{std}.
% For each table, we use horizontal lines to partition the table into various levels of sub-tables.
% Each sub-table contains results for methods that have comparable settings (\eg architecture).
% That is, \texttt{Result1} should only be compared to \texttt{Result2} if there is no horizontal line separating them in the table.
% \aaron{Check that all tables follow this}
% In each sub-table, we highlight the best performing method in \colorbox{lred}{red} and the second-best performing method in \colorbox{lblue}{blue}.
% If there are only two comparable methods, then we just highlight the best performing method.
% , and the third-best performing method in \colorbox{lyellow}{yellow}.

\subsection{\methodsp Details}
\paragraph{\methodsp Variants}
We consider three~\methodsp variants, each with a different combination of weights $\lambda_{\text{KD-In}}$ and $\lambda_{\text{KD-Out}}$ (\textsection \ref{sec:method:kd}).
\textbf{\methodsp (In)} trains the student LM with $\mathcal{L}_{\text{KD-In}}=1$ and $\mathcal{L}_{\text{KD-Out}}=0$.
\textbf{\methodsp (Out)} trains the student LM with $\mathcal{L}_{\text{KD-Out}}=1$ and $\mathcal{L}_{\text{KD-In}}=0$.
\textbf{\methodsp (In+Out)} trains the student LM with $\mathcal{L}_{\text{KD-In}}=\mathcal{L}_{\text{KD-Out}}=1$.
By default, we use~\methodsp (In+Out).

% As an upper bound, Tables \ref{tab:exp:abl:ftr-quality}-\ref{tab:exp:abl:teacher-bneck} report the performance of the \textbf{\methodsp Teacher}, which has access to FTRs during both training and inference.

\paragraph{Implementation Details}
Following prior works \citep{narang2020wt5, sun2022investigating, wiegreffe2021measuring}, we use T5-Base and T5-Large \citep{raffel2020exploring} as the backbone model for~\methodsp and all baselines.
For KD-based methods, \textbf{T5-Base$\rightarrow$T5-Base} means teacher and student use T5-Base, \textbf{T5-Large$\rightarrow$T5-Large} means teacher and student use T5-Large, and \textbf{T5-Large$\rightarrow$T5-Base} means T5-Large teacher and T5-Base student.\footnote{For T5-Large$\rightarrow$T5-Base, the teacher and student have different representation spaces of hidden states.
Thus, in KD we jointly train two linear projection layers to transform the student LM's encoder and decoder hidden states to be in the same representation space as the teacher LM's.}
For non-KD methods used for comparison with the three ones, they use T5-Base, T5-Large, and T5-Base, respectively.
% For \method, the student uses the teacher's language modeling head.
In our implementation of~\method, if the student and teacher have the backbone model (\ie both T5-Base or both T5-Large), the student's parameters are initialized as the teacher's. 
% \zhiyuan{Actually for all variants the student LM uses the teacher LM's language modeling head, not just for large2base. This is because, T5 takes the embedding layer's parameters as LM head's parameters, so student LM's LM head is actually changed after fine-tuning. Besides, is the term "decoder output head" common?}
% \paragraph{Hyperparameters}
% For the KD, we used sweeps of $\lambda_{\text{task}} = [0, 1]$, $\lambda_{\text{KD-in}} = [0, 1]$, and $\lambda_{\text{KD-out}} = [0, 1]$.
\textsection \ref{sec:app:hparams} lists the hyperparameters.

\subsection{Baselines}
\label{sec:exp:baseline}
We consider a wide range of baselines, spanning various existing paradigms.

\textbf{Standard Finetuning} does not involve FTRs or KD.
Without any use of FTRs, \textbf{FT (I$\rightarrow$O)} finetunes a T5 on the task dataset by the cross-entropy loss.

\textbf{Finetuned Self-Rationalization} inserts an FTR to the LM's target output during training.
\textbf{FT (I$\rightarrow$OR)} finetunes a T5 to generate the task output followed by the FTR.
\textbf{FT (I$\rightarrow$RO)} finetunes a T5 to generate the FTR followed by the task output.
They both take as input the raw task input.\footnote{As labels other than the gold label don't have FTRs, FT (I$\rightarrow$OR) and FT (I$\rightarrow$RO) cannot use the basic design in \textsection \ref{sec:method:lm}. Instead, we train the LM by teacher-forcing the target output to the decoder. During inference, we use greedy decoding to generate the prediction.}
%\footnote{FT (I$\rightarrow$OR) and FT (I$\rightarrow$RO) use greedy decoding (\textsection \ref{sec:method:lm}) because, unlike task labels, there do not exist multiple FTR choices (besides gold FTR $\mathbf{y}^*$) to separately teacher-force into the decoder.}

\textbf{Input Augmentation} appends an FTR to the LM's input during training.
\textbf{FT (IR$\rightarrow$O)} finetunes a T5 to predict the output taking as input the task input concatenated with the corresponding FTR.
% However, for a fair comparison, the FTR input is omitted during inference.
This is equivalent to training~\method's teacher LM without the bottleneck design.
As the inference instances do not have FTRs, the input during inference is just the raw task input.
The global patterns that FTR is appended to the training input are absent during inference, which causes the issue of input distribution shift.
\textbf{FT Dropout (IR$\rightarrow$O)} randomly drops out the appended FTR during training, in order to mitigate input distribution shift by also training the model to deal with the raw task input.

\textbf{Prompted Self-Rationalization} uses chain-of-thought (CoT) prompting \citep{wei2022chain}. \textbf{CoT (I$\rightarrow$RO)} prompts GPT-NeoX and GPT-3 (text-davinci-003) to generate the FTR followed by the task output by CoT prompting.\footnote{All CoT (I$\rightarrow$RO) results were obtained from \citet{wang2022pinto}, except GPT-3 on OBQA, which was obtained from \citet{huang2022large}.}

\textbf{Pipeline Rationalization} finetunes two LMs as a pipeline.
For \textbf{FT (I$\rightarrow$R$\rightarrow$O)}, the first T5 is finetuned to generate the FTR given the task input, while the second T5 is finetuned to generate the task output given the first T5's generated FTR.\footnote{Since this method is known to not perform well~\citep{wiegreffe2021measuring, wang2022pinto}, we only consider FT (I$\rightarrow$R$\rightarrow$O) in a limited set of settings. In these settings, we present the results reported in~\citet{wang2022pinto}. Note that only the mean performance is available.}

\textbf{FT Teacher Init.} finetunes a T5 as FT (I$\rightarrow$O) does, with initializing the LM with the \methodsp teacher's parameters.
Intuitively, the teacher's parameters store reasoning knowledge, so it is natural to view parameter initialization as a way of transferring such knowledge.

\begin{wraptable}{R}{0.6\textwidth}
\vspace{-0.2cm}
\centering
\scalebox{0.63}{
\begin{tabular}{cccc}
    \toprule
    \multirow{2}{*}{\textbf{Architecture}} & \multirow{2}{*}{\textbf{Method}} & \multicolumn{2}{c}{\textbf{Accuracy} ($\uparrow$)} \\
    % \cmidrule(lr){3-5}
    \cmidrule(lr){3-4}
    % & & OBQA & StrategyQA & QuaRTz \\
    & & OBQA & StrategyQA \\
    
    \midrule
    
    \multirow{7}{*}{T5-Base$\rightarrow$T5-Base} & FT (I$\rightarrow$O) & 57.93~($\pm$1.15) & 59.05~($\pm$0.23) \\ % & 68.20~($\pm$0.52) \\
    
    & FT (I$\rightarrow$OR) & 53.93~($\pm$1.33) & 51.84~($\pm$1.45) \\ %& 57.19~($\pm$0.58) \\
    
    & FT (I$\rightarrow$RO) & 55.53~($\pm$0.46) & \cellcolor{lblue} 58.65~($\pm$1.53) \\ %& 29.38~($\pm$1.85) \\

    & FT (I$\rightarrow$R$\rightarrow$O) & 56.65 & 57.11 \\ %& - \\

    & FT (IR$\rightarrow$O) & 53.73~($\pm$2.31) & 49.97~($\pm$2.92) \\ %& 66.41~($\pm$0.90) \\

    & FT Dropout (IR$\rightarrow$O) & 58.27~($\pm$1.33) & 55.85~($\pm$2.09) \\ %& 68.75~($\pm$0.34) \\

    & FT Teacher Init. & \cellcolor{lblue} 58.33~($\pm$0.90) & 57.25~($\pm$2.22) \\ %& 68.20~($\pm$0.52) \\
    
    % & \methodsp (In) & 60.00~($\pm$0.40) & \cellcolor{lred} 61.39~($\pm$1.90) \\ %  & \textbf{68.45}~($\pm$0.83) \\
    
    % & \methodsp (Out) & \cellcolor{lred} 62.27~($\pm$1.01) & 59.05~($\pm$1.22) \\ %  & \textbf{68.45}~($\pm$0.52) \\
    
    & \methodsp (In+Out) & \cellcolor{lred} 61.53~($\pm$0.76) & \cellcolor{lred} 60.45~($\pm$0.31) \\ %& 68.41~($\pm$0.99) \\

    % \cmidrule(lr){2-5}

    % & CoT Prompting (GPT-3) & I$\rightarrow$RO & X.XX~($\pm$X.XX) & X.XX~($\pm$X.XX) & X.XX~($\pm$X.XX) \\
    
    % & \methodsp Teacher Finetuning & IR$\rightarrow$O & 73.80~($\pm$0.60) & 66.20~($\pm$1.10) \\ %  & 75.85~($\pm$2.20) \\

    \midrule
    
    \multirow{6}{*}{T5-Large$\rightarrow$T5-Large} & FT (I$\rightarrow$O) & 65.60~($\pm$0.40) & 57.58~($\pm$0.70) \\ %& 74.11~($\pm$1.93) \\
    
    & FT (I$\rightarrow$OR) & 61.93~($\pm$1.97) & 57.58~($\pm$0.12) \\ %& 65.65~($\pm$1.24) \\
    
    & FT (I$\rightarrow$RO) & 61.87~($\pm$2.12) & \cellcolor{lblue} 63.66~($\pm$1.14) \\ %& 35.80~($\pm$1.60) \\

    & FT (IR$\rightarrow$O) & 61.27~($\pm$2.16) & 53.24~($\pm$2.54) \\ %& 70.49~($\pm$0.77) \\

    & FT Dropout (IR$\rightarrow$O) & \cellcolor{lblue} 65.73~($\pm$1.36) & 59.25~($\pm$4.59) \\ %& 74.74~($\pm$0.67) \\

    & FT Teacher Init. & 65.67~($\pm$2.25) & 61.72~($\pm$2.36) \\ %& 74.83~($\pm$1.16) \\
    
    % & \methodsp (In) & 66.20~($\pm$0.53) & 62.66~($\pm$3.38) \\ %  & 72.41~($\pm$0.39) \\
    
    % & \methodsp (Out) & \cellcolor{lblue} 68.07~($\pm$1.50) & \cellcolor{lred} 64.40~($\pm$1.22) \\ %  & \textbf{74.70}~($\pm$1.81) \\
    
    & \methodsp (In+Out) & \cellcolor{lred} 68.73~($\pm$1.36) & \cellcolor{lred} 63.79~($\pm$0.64) \\ %& 73.89~($\pm$1.60) \\

    % \cmidrule(lr){2-5}

    % % & CoT Prompting (GPT-3) & I$\rightarrow$RO & X.XX~($\pm$X.XX) & X.XX~($\pm$X.XX) & X.XX~($\pm$X.XX) \\
    
    % & \methodsp Teacher Finetuning & IR$\rightarrow$O & 80.93~($\pm$0.12) & 81.83~($\pm$1.16) \\ %  & 88.05~($\pm$0.90) \\

    \midrule
    
    % \multirow{10}{*}{T5-Large} & FT (I$\rightarrow$O) & I$\rightarrow$O & 57.93~($\pm$1.15) & 59.05~($\pm$0.23) \\ %  & 67.05~($\pm$0.78) \\
    
    % \multirow{10}{*}{$\downarrow$} & FT (IR$\rightarrow$O) & IR$\rightarrow$O & 53.73~($\pm$2.31) & 49.97~($\pm$2.92) \\ %  & 66.41~($\pm$0.90) \\

    % \multirow{10}{*}{T5-Base} & FT Dropout (IR$\rightarrow$O) & IR$\rightarrow$O & 58.27~($\pm$1.33) & 59.39~($\pm$1.33) \\ %  & 66.41~($\pm$0.90) \\
    
    % & FT & I$\rightarrow$OR & 53.93~($\pm$1.33) & 51.84~($\pm$1.45) \\ %  & 57.19~($\pm$0.58) \\
    
    % & FT & I$\rightarrow$RO & 39.40~($\pm$1.20) & 0.87~($\pm$0.50) \\ %  & 29.38~($\pm$1.85) \\

    \multirow{2}{*}{T5-Large$\rightarrow$T5-Base} & Best T5-Base$\rightarrow$T5-Base & \cellcolor{lblue} 58.33~($\pm$0.90) & \cellcolor{lblue} \cellcolor{lblue} 58.65~($\pm$1.53) \\ %& 68.75~($\pm$0.34) \\

    % \multirow{6}{*}{$\downarrow$} & Non-FTR KD (In) \zhiyuan{Explain} & 31.87~($\pm$2.10) & 53.77~($\pm$0.46) \\

    % \multirow{6}{*}{T5-Base} & Non-FTR KD (Out) & 55.60~($\pm$2.99) & 59.99~($\pm$3.53) \\ %  & 65.43~($\pm$1.22) \\
    
    % & Non-FTR KD (In+Out) & 58.27~($\pm$1.01) & \cellcolor{lblue} 60.12~($\pm$1.40) \\ %  & \textbf{66.75}~($\pm$0.19) \\

    % & Non-FTR KD (Logit) & 48.53~($\pm$4.06) & 57.72~($\pm$2.12) \\ %  & \textbf{66.75}~($\pm$0.19) \\

    % & \methodsp (In) \zhiyuan{Explain} & 31.13~($\pm$2.87) & 53.77~($\pm$0.46) \\ %  & 65.43~($\pm$1.22) \\
    
    % & \methodsp (Out) & 55.60~($\pm$2.42) & \cellcolor{lred} 61.12~($\pm$0.53) \\ %  & 65.43~($\pm$1.22) \\
    
    & \methodsp (In+Out) & \cellcolor{lred} 60.93~($\pm$0.12) & \cellcolor{lred} 61.12~($\pm$2.03) \\ %& 66.75~($\pm$0.19) \\

    % \cmidrule(lr){2-5}

    % % & CoT Prompting (GPT-3) & I$\rightarrow$RO & X.XX~($\pm$X.XX) & X.XX~($\pm$X.XX) & X.XX~($\pm$X.XX) \\
    
    % & \methodsp Teacher Finetuning & IR$\rightarrow$O & 80.93~($\pm$0.12) & 81.83~($\pm$1.16) \\ %  & 88.05~($\pm$0.90) \\

    \midrule

    GPT-NeoX & CoT (I$\rightarrow$RO) & 33.80 & 55.31 \\

    % & PINTO & IR$\rightarrow$O & 58.85 & 60.87 \\

    \midrule
    
    GPT-3 (text-davinci-003) & CoT (I$\rightarrow$RO) & 86.40 & 66.53 \\
    
    \bottomrule 
\end{tabular}
}
% \vspace{0.3cm}
\caption{\small \textbf{\methodsp Main Results}
}
\label{tab:exp:main:gold}
\vspace{-0.5cm}
\end{wraptable}

\subsection{Main Results}
\label{sec:exp:main}

Table \ref{tab:exp:main:gold} presents our main results.
Here, LMs are finetuned on the entire training set.
Methods requiring FTRs use the human-annotated FTRs (called gold FTRs in the following paragraphs) provided by the public datasets.
% We describe our findings below.
We observe that \methodsp (In+Out) consistently outperforms all baselines, suggesting \methodsp effectively extracts reasoning knowledge from FTRs of training instances and utilizes such knowledge for performance improvement.
Besides, \methodsp is the only FTR-requiring method that consistently outperforms FT (I$\rightarrow$O), which shows the difficulty of improving small models' task performance by FTRs.

We observe that FT (I$\rightarrow$OR) and FT (I$\rightarrow$RO) often bring performance drop compared to FT (I$\rightarrow$O), which is consistent with observations from prior works~\citep{hase2021can, zhang2023multi}.
Our explanation is that task and FTR generation objectives could conflict with each other, meaning that jointly optimizing them may hurt task performance.
Besides, the generated FTRs could contain hallucinated information misleading the answer prediction~\citep{zhang2023multi}.

FT (IR$\rightarrow$O) is significantly worse than FT (I$\rightarrow$O), which is expected due to the input distribution shift issue (\textsection \ref{sec:exp:baseline}).
FT Dropout (IR$\rightarrow$O) mitigates the issue to some extent but still fails to bring consistent improvement.
The results also support the argument that we cannot directly use \method's teacher LM during inference (\textsection \ref{sec:method:lm}), which is basically FT (IR$\rightarrow$O) with bottleneck design.

FT (I$\rightarrow$R$\rightarrow$O) is also worse than FT (I$\rightarrow$O).
It is because the generated FTR could be poor, omitting critical information from the task input or introducing irrelevant and even incorrect information, which can hurt task performance~\citep{huang2022large, magister2022teaching, li2022explanations}.
% Unlike FT (I$\rightarrow$R$\rightarrow$O), \methodsp (In+Out)'s LM has direct access to the task input when predicting the task output\footnote{Note that only the mean performance is available for FT (I$\rightarrow$R$\rightarrow$O) since these results were obtained from \citet{wang2022pinto}.}.
% Unlike FT (IR$\rightarrow$O) and FT Dropout (I$\rightarrow$RO), \methodsp (In+Out) avoids input distribution shift since its student only takes task input in training and inference.
% Unlike FT Teacher Init. (I$\rightarrow$O), \methodsp (In+Out) explicitly trains the student to follow the teacher's reasoning process.
% Plus, the fact that FT Teacher Init. (I$\rightarrow$O)'s initial parameters were optimized for FTR-augmented input may also introduce an input distribution shift.  
% This may be due to the projection layer causing information loss when transferring knowledge between different architectures. \zhiyuan{How to improve that is left for future work. It is meaningful because generally, we hope large teachers can provide more information.}

FT Teacher Init. outperforms FT (I$\rightarrow$O) in most cases, but its improvement is much smaller than that of \methodsp (In+Out).
Thus, parameter initialization is a potential way to transfer reasoning knowledge from the teacher to the student, but it is much less effective than knowledge distillation.

Also, we report CoT (I$\rightarrow$RO) results for GPT-NeoX (20B) \citep{black2022gpt} and GPT-3 (text-davinci-003, 175B) \citep{brown2020language}.
% \zhiyuan{How about adding this sentence to footnote?}
Since GPT-NeoX and GPT-3 are much larger than T5-Base (220M) and T5-Large (770M), it is unfair to expect other methods to perform as well as CoT (I$\rightarrow$RO).
Even so, we find that \methodsp (In+Out) greatly outperforms GPT-NeoX on all settings, while \methodsp (In+Out) with T5-Large achieves similar performance to GPT-3 on StrategyQA.

In Table~\ref{tab:exp:main:gpt}, we repeat these experiments for GPT-NeoX generated FTRs~\citep{wang2022pinto} and find that~\methodsp (In+Out) still consistently outperforms all baselines.
It shows~\method's robustness to different sources of FTRs.
% This makes sense since GPT-NeoX's FTRs are known to fluently convey useful knowledge \citep{black2022gpt}.
Interestingly,~\methodsp with GPT-NeoX FTRs still considerably outperforms CoT (I$\rightarrow$RO) with GPT-NeoX, despite~\methodsp using much smaller LMs.
% This demonstrates the effectiveness of \method's KD training process.

In Table~\ref{tab:exp:low}, we consider a low-resource setting where available is only 10\% of the training data, using T5-Base$\rightarrow$T5-Base on OBQA.
We find~\methodsp beats all baselines, showing that~\methodsp effectively extracts and injects reasoning knowledge also in the low-resource setting.

% \zhiyuan{Quickly mention low-resource setting here?}

% Fourth, as a sanity check, we see that \methodsp Teacher Finetuning outperforms all other methods.
% Note that \methodsp Teacher Finetuning is not directly comparable \zhiyuan{We should emphasize this point in the table.} to other methods because it has access to FTRs during both training and inference.
% Also, despite its FTR access, \methodsp Teacher Finetuning's performance is still relatively far from perfect.
% This suggests that the FTRs are not label-leaking \zhiyuan{This discussion can be moved to Case Study. Besides, I think the discussion on label-leaking is not very interesting.}. 

% Fifth, as expected, we find that T5-Large based methods generally outperform their T5-Base counterparts.
% However, although \methodsp Student Finetuning outperforms Vanilla Finetuning for T5-Large $\rightarrow$ T5-Base, we find that T5-Large $\rightarrow$ T5-Base does not outperform T5-Base for \methodsp Student Finetuning.
% This may be due to the linear projection layer causing information loss when transferring knowledge between the two different architectures. \zhiyuan{How to improve that is left for future work. It is meaningful because generally, we hope large teachers can provide more information.}

\begin{wraptable}{R}{0.6\textwidth}
\vspace{-1.1cm}
\centering
\scalebox{0.6}{
\begin{tabular}{ccccc}
    \toprule
    \multirow{2}{*}{\textbf{Architecture}} & \multirow{2}{*}{\textbf{Method}} & \multicolumn{2}{c}{\textbf{Accuracy} ($\uparrow$)} \\
    \cmidrule(lr){3-4}
    & & OBQA & StrategyQA \\
    
    \midrule
    
    \multirow{6}{*}{T5-Base$\rightarrow$T5-Base} & \methodsp (In) + Gold & 60.00~($\pm$0.40) & \cellcolor{lred} 61.39~($\pm$1.90) \\ %  & \textbf{68.45}~($\pm$0.83) \\
    
    & \methodsp (Out) + Gold & \cellcolor{lred} 62.27~($\pm$1.01) & 59.05~($\pm$1.22) \\ %  & \textbf{68.45}~($\pm$0.52) \\
    
    & \methodsp (In+Out) + Gold & \cellcolor{lblue} 61.53~($\pm$0.76) & 
    \cellcolor{lblue} 60.45~($\pm$0.31) \\ %  & 68.41~($\pm$0.99) \\

    \cmidrule{2-4}

    & \methodsp (In) + GPT-NeoX & 61.07~($\pm$0.12) & \cellcolor{lred} 61.92~($\pm$1.74) \\
    
    & \methodsp (Out) + GPT-NeoX & \cellcolor{lred} 61.60~($\pm$0.53) & \cellcolor{lblue} 60.72~($\pm$0.20) \\
    
    & \methodsp (In+Out) + GPT-NeoX & \cellcolor{lblue} 61.53~($\pm$0.76) & \cellcolor{lred} 61.92~($\pm$1.04) \\

    \midrule
    
    \multirow{6}{*}{T5-Large$\rightarrow$T5-Large} & \methodsp (In) + Gold & 66.20~($\pm$0.53) & 62.66~($\pm$3.38) \\ %  & 72.41~($\pm$0.39) \\
    
    & \methodsp (Out) + Gold & \cellcolor{lblue} 68.07~($\pm$1.50) & \cellcolor{lred} 64.40~($\pm$1.22) \\ %  & \textbf{74.70}~($\pm$1.81) \\
    
    & \methodsp (In+Out) + Gold & \cellcolor{lred} 68.73~($\pm$1.36) & \cellcolor{lblue} 63.79~($\pm$0.64) \\ %  & 73.89~($\pm$1.60) \\

    \cmidrule{2-4}

    & \methodsp (In) + GPT-NeoX & 67.20~($\pm$0.40) & \cellcolor{lblue} 62.32~($\pm$1.84) \\
    
    & \methodsp (Out) + GPT-NeoX & \cellcolor{lblue} 68.53~($\pm$1.89) & 62.26~($\pm$0.64) \\
    
    & \methodsp (In+Out) + GPT-NeoX & \cellcolor{lred} 68.73~($\pm$1.55) & \cellcolor{lred} 63.99~($\pm$0.81) \\

    \midrule

    \multirow{6}{*}{T5-Large$\rightarrow$T5-Base} & \methodsp (In) + Gold & 31.13~($\pm$2.87) & \cellcolor{lblue} 53.77~($\pm$0.46) \\ %  & 65.43~($\pm$1.22) \\
    
    & \methodsp (Out) + Gold & \cellcolor{lblue} 55.60~($\pm$2.42) & \cellcolor{lred} 61.12~($\pm$0.53) \\ %  & 65.43~($\pm$1.22) \\
    
    & \methodsp (In+Out) + Gold & \cellcolor{lred} 60.93~($\pm$0.12) & \cellcolor{lred} 61.12~($\pm$2.03) \\ %  & \textbf{66.75}~($\pm$0.19) \\

    \cmidrule{2-4}

    & \methodsp (In) + GPT-NeoX & 30.07~($\pm$2.97) & 53.91~($\pm$0.69) \\
    
    & \methodsp (Out) + GPT-NeoX & \cellcolor{lblue} 55.60~($\pm$2.03) & \cellcolor{lblue} 60.59~($\pm$0.61) \\
    
    & \methodsp (In+Out) + GPT-NeoX & \cellcolor{lred} 60.47~($\pm$0.81) & \cellcolor{lred} 62.39~($\pm$0.42) \\
    
    \bottomrule 
\end{tabular}
}
% \vspace{0.3cm}
\caption{\small \textbf{\methodsp Variants}
}
\label{tab:exp:abl:knife-variants}
\vspace{-0.7cm}
\end{wraptable}

\subsection{Ablation Studies}
\label{sec:exp:abl}

% To validate our \methodsp design choices, w
To justify design choices of~\methodsp and understand why it works, we present ablation studies, analyzing the impacts of KD objective, FTR usage, FTR perturbation, teacher bottleneck, and student task loss.
% on \method's performance.
% \zhiyuan{To meet the page limit requirement, we may move some ablation studies to the Appendix. From my perspective, the importance order of the three ablation studies is Bottleneck>Task Loss>FTR type. This could be a reference when considering moving which ones to the Appendix.}

% \zhiyuan{I hope we can add a footnote on large2base KNIFE Student Finetuning (IN) to explain it should not work.}

\paragraph{KD Objectives}
Table \ref{tab:exp:abl:knife-variants} compares the performance of~\methodsp (In),~\methodsp (Out), and~\methodsp (In+Out).
For both gold (human-annotated) and GPT-NeoX (model-generated) FTRs, we find that~\methodsp (In+Out) generally achieves the highest performance with a few exceptions.
This suggests useful FTR knowledge can be distilled via both encode and decoder hidden states, so it is recommended to use both by default.
Furthermore,~\methodsp (In) and~\methodsp (Out) are also able to outperform all baselines in almost all cases.

% Although \methodsp (In) and \methodsp (Out) perform similarly for T5-Base$\rightarrow$T5-Base and T5-Large$\rightarrow$T5-Large, 
\methodsp (In) performs much worse for T5-Large$\rightarrow$T5-Base than the two others and many baselines.
Since~\methodsp (In) only performs KD via the encoder hidden states, the student's decoder is not trained.
Here, the student cannot be initialized with the teacher's parameters as their backbone models are different, leaving the student's decoder with T5-Base's pretrained parameters.
Thus, in such cases,~\methodsp (In) is problematic and necessary is KD via the decoder hidden states.
% While this could be addressed by training the student with task loss, it shows a major disadvantage of \methodsp (In) compared to other \methodsp variants.

\begin{figure*}[h!]
% \vspace{-0.5cm}
	\begin{subfigure}[b]{0.25\linewidth}
		\centering
		\includegraphics[width=\textwidth]{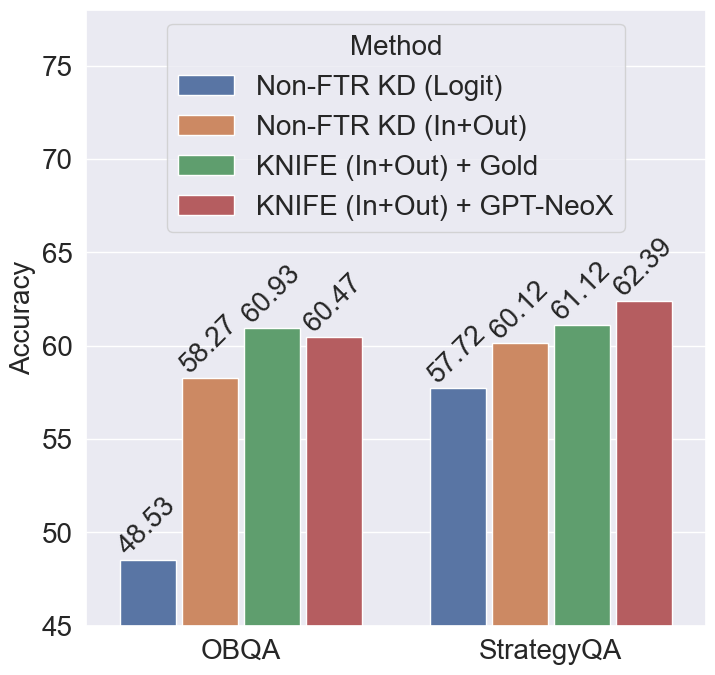}
		\caption{FTR Usage}
		\label{fig:exp:abl:ftr-usage}
	\end{subfigure}\hfill
	\begin{subfigure}[b]{0.25\linewidth}
		\centering
		\includegraphics[width=\textwidth]{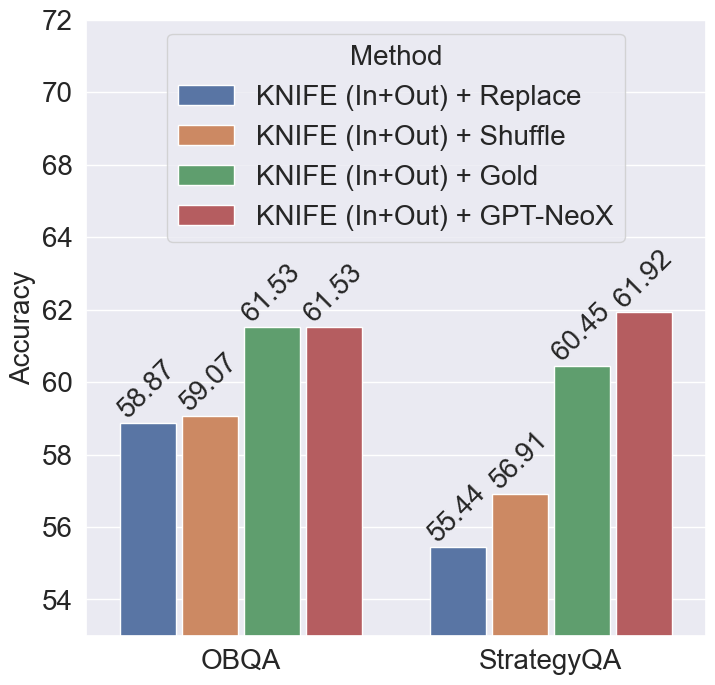}
		\caption{FTR Perturbation}
		\label{fig:exp:abl:ftr-quality}
	\end{subfigure}\hfill
	\begin{subfigure}[b]{0.25\linewidth}
		\centering
		\includegraphics[width=\textwidth]{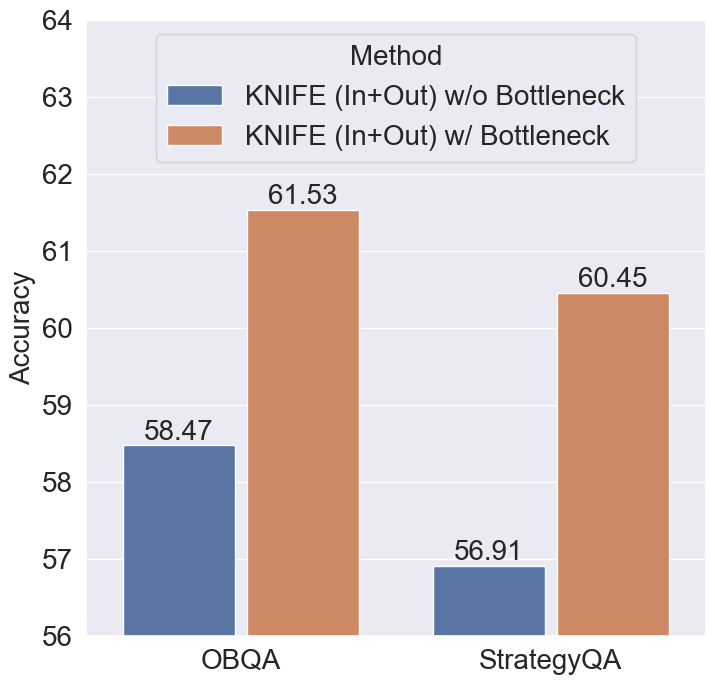}
		\caption{Teacher Bottleneck}
		\label{fig:exp:abl:teacher-bneck}
	\end{subfigure}\hfill
        \begin{subfigure}[b]{0.25\linewidth}
		\centering
		\includegraphics[width=\textwidth]{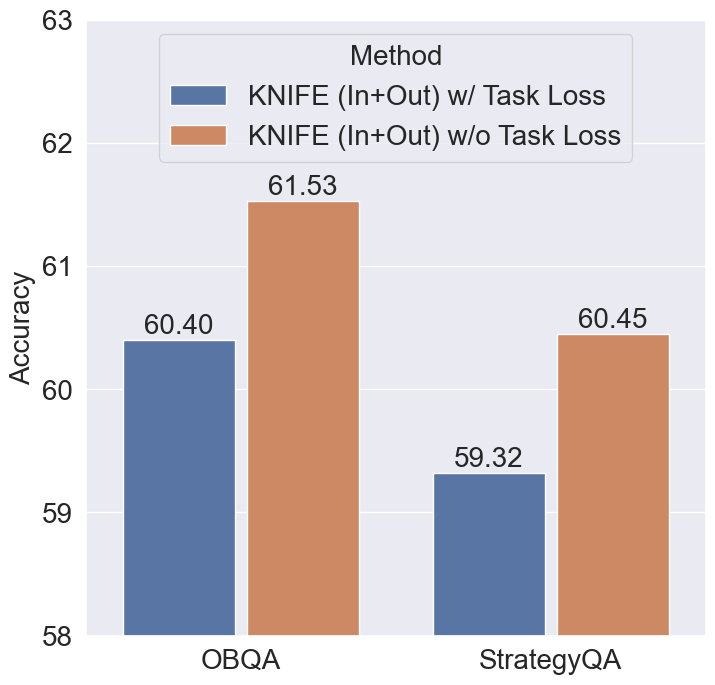}
		\caption{Student Task Loss}
		\label{fig:exp:abl:student-taskloss}
	\end{subfigure}
        
	\caption{\small \textbf{\methodsp Ablation Studies}}
	\label{fig:exp:abl}
    \vspace{-0.3cm}
\end{figure*}

\paragraph{FTR Usage}
KD has been widely used to transfer knowledge from a larger teacher to a smaller student~\citep[etc]{hinton2015distilling, sanh2019distilbert}, so the larger teacher's capacity could be one source of performance gain for T5-Large$\rightarrow$T5-Base.
To verify FTR usage's importance for the performance gain, we compare~\methodsp to non-FTR KD methods, where the teacher is FT (I$\rightarrow$O).
% This experiment verifies the importance of FTR usage in KD by comparing \methodsp to non-FTR KD baselines, where the teacher is FT (I$\rightarrow$O).
\textbf{Non-FTR KD (Logit)} trains the student to align its logit distribution with the teacher's.
% \textbf{Non-FTR KD (In)} finetunes the student so its task input states align with the teacher's.
% \aaron{Make sure we have the Non-FTR KD (In) results}
% \textbf{Non-FTR KD (Out)} finetunes the student so its task output states aligns with the teacher's.
\textbf{Non-FTR KD (In+Out)} trains the student to align its encoder and decoder hidden states with the teacher's.
% Since non-FTR KD generally requires the teacher to be larger than the student, we only use T5-Large$\rightarrow$T5-Base here.
In Fig. \ref{fig:exp:abl:ftr-usage}, both~\methodsp (In+Out) + Gold and~\methodsp (In+Out) + GPT-NeoX outperform the non-FTR KD baselines, showing that~\methodsp actually benefits from FTRs and thus the reasoning knowledge.
Plus, Non-FTR KD (Logit) performs much worse than Non-FTR KD (In+Out), which validates~\method's use of representation-based KD.

\paragraph{FTR Perturbation}
% By default, \methodsp uses gold FTRs to train the teacher, and we have also explored using GPT-NeoX FTRs.
% Since gold FTRs are human-annotated to support the gold label, we can assume that gold FTRs contain meaningful knowledge for reasoning about the given downstream task.
% Using T5-Base$\rightarrow$T5-Base, we investigate the relationship between \methodsp performance and FTR quality by considering \methodsp variants that train the teacher on noisy FTRs:
To prove the performance gain of~\methodsp is not from any unexpected noise, \eg more computation of the teacher LM due to longer input, we perturb the FTRs and observe how it influences the performance of ~\method.
\textbf{Replace} replaces each FTR token with a random token from the token vocabulary.
\textbf{Shuffle} shuffles FTRs across all training instances in the dataset.
% , so that each training instance is paired with a gold FTR from a different training instance.
% We also compare to .
% \textbf{Gold} denotes the default \methodsp variant that uses gold FTRs to train the teacher LM.
% \zhiyuan{Teacher Finetuning in the table is not very meaningful (especially compared to Student Finetuning), I think we should try to deemphasize it in the table to make readers pay much more attention to Student Finetuning.}
In Fig. \ref{fig:exp:abl:ftr-quality}, \methodsp (In+Out)'s performance with Replace and Shuffle is much lower than with Gold and GPT-NeoX, which shows the performance gain actually comes from the FTR knowledge.
% Plus, Replace performs worse than Shuffle since Replace's token-level randomization fully corrupts the FTR.
% This suggests that \method's performance is positively correlated with FTR quality.
% , so \methodsp cannot work unless its FTRs contain meaningful knowledge.

% In this experiment, we consider various datasets (OBQA, StrategyQA) and architectures (T5-Base, T5-Large, T5-Large $\rightarrow$ T5-Base). For \methodsp configurations, we consider Student Finetuning (In), Student Finetuning (Out), Student Finetuning (In+Out), and Teacher Finetuning.
% As expected, we observe that Gold consistently outperforms Replace and Shuffle, often by significant margins.
% This suggests that FTR type/quality matters a lot and that gold FTRs' knowledge is indeed the source of \method's performance improvements over the Vanilla Finetuning baselines.
% However, on some settings like OBQA + T5-Large $\rightarrow$ T5-Base, Gold is not quite as dominant.
% In particular, for Student Finetuning (Out) on OBQA + T5-Large $\rightarrow$ T5-Base, both Replace and Shuffle slightly outperform Gold \zhiyuan{In these settings, the Student Finetuning (Out) cannot beat baselines, so actually we do not need to compare them with Replace and Shuffle. We can omit this discussion.}.
% As we saw in the main results, this may again be due to the linear projection layer causing information loss when transferring knowledge between the two different architectures, so that FTR quality matter less.

\paragraph{Teacher Bottleneck}
% Previously, we asserted that the teacher LM's FTR bottleneck is critical for distilling knowledge from the teacher LM's FTR states to its task input/output states, through which FTR knowledge is ultimately distilled to the student LM.
We empirically validate the bottleneck's necessity for KD via encoder hidden states by training a teacher without the bottleneck.
% We validate \method's teacher bottleneck by considering a \methodsp variant with no bottleneck.
% Specifically, given a teacher LM with no FTR bottleneck, we report the performance of both the teacher LM and its corresponding student LM.
In Fig.~\ref{fig:exp:abl:teacher-bneck} and Table~\ref{tab:exp:abl:ftr-bneck}, we show that removing bottleneck brings significant performance drop to \methodsp (In+Out) and \methodsp (In), but it is not as critical for \methodsp (Out).
It is expected as the bottleneck of~\method's teacher is designed to address the issue that the student and teacher have different structures of encoder hidden states, and KD via only decoder hidden states doesn't meet the issue.
% \zhiyuan{Teacher Finetuning in the table is not very meaningful (especially compared to Student Finetuning), I think we should try to deemphasize it in the table to make readers pay much more attention to Student Finetuning.} shows the results of this ablation study.
% In this experiment, we consider the OBQA and StrategyQA datasets, the T5-Base and T5-Large architectures, and Student Finetuning (In) as a representative \methodsp configuration.
% For Teacher Finetuning, there seems to be little correlation between task performance and the use of the FTR bottleneck.
% This demonstrates the bottleneck's effectiveness in routing FTR knowledge to the teacher's task input/output states, through which FTR knowledge can be distilled to the student.
% \zhiyuan{In : vitally important   In+Out : very important  Out : not so important relatively}
% , so that the teacher LM can still leverage FTR context for solving the task even when the FTR states are masked out.
% For Student Finetuning (In), we observe that having the FTR bottleneck always improves task performance, especially for StrategyQA.
% This supports our hypothesis that, since the teacher LM's task input/output states are well-infused with FTR knowledge, it is also effective to distill knowledge from the teacher LM's task input/output states to the student LM's.

\paragraph{Student Task Loss}
By default,~\methodsp trains the student with only KD losses.
We justify it by comparing it to~\methodsp variants where the task loss is also involved in the total loss.
Specifically,~\methodsp with task loss trains the student by the loss $\mathcal{L} = \lambda_{\text{KD-In}} \mathcal{L}_{\text{KD-In}} + \lambda_{\text{KD-Out}} \mathcal{L}_{\text{KD-Out}} + \mathcal{L}_{\text{task}}$.
In Fig.~\ref{fig:exp:abl:student-taskloss} and Table~\ref{tab:exp:abl:task-loss}, we see omitting the task loss consistently yields higher performance.
%, which indicates that KD losses are not always compatible with task loss.
Intuitively, KD loss is guided by the supervision signal of reasoning knowledge, while task loss is guided by the supervision signal of task labels, which could be sub-optimal for learning reasoning knowledge.
Thus, the KD and task losses could conflict during optimization, which hurts the performance.
% displays the results for the ablation study.
% We consider the OBQA and Strategy QA datasets as well as the T5-Base architecture.
% Also, we consider the Student Finetuning (In), Student Finetuning (Out), and Student Finetuning (In+Out) \methodsp configurations.
% Across all settings, we see that omitting the task loss always yields better performance.
% This may be due to the teacher LM sometimes exhibiting reasoning processes (KD loss) that conflict with the gold label (task loss).
% \zhiyuan{This explanation is not convincing to me. In my understanding, the reason could be that during optimization two losses' gradients have conflicts.}.
% If the KD and task losses produce conflicting gradients during optimization, the student may get confused and learn a suboptimal ``in-between'' reasoning process leading to even worse generalization.

% \subsection{Case Study \zhiyuan{Rethink the title}}
\subsection{Failure Analysis}
\label{sec:exp:failure}

Although \methodsp performs well on OBQA and StrategyQA, it yields negative results on other QA datasets like ECQA \citep{ecqa} and QuaRTz \citep{quartz}.
Using T5-Base$\rightarrow$T5-Base, we compare the performance of \methodsp and the main baselines considered in the main results.
In Table \ref{tab:exp:negative}, we see that \methodsp generally outperforms all FTR-based baselines, sometimes by a very large margin.
Still, none of the FTR-based methods (including all \methodsp variants) are able to significantly outperform FT (I$\rightarrow$O).

\begin{wraptable}{R}{0.55\textwidth}
\vspace{-0.7cm}
\centering
\scalebox{0.68}{
\begin{tabular}{cccc}
    \toprule
    \multirow{2}{*}{\textbf{Architecture}} & \multirow{2}{*}{\textbf{Method}} & \multicolumn{2}{c}{\textbf{Accuracy} ($\uparrow$)} \\
    % \cmidrule(lr){3-5}
    \cmidrule(lr){3-4}
    & & ECQA & QuaRTz \\
    
    \midrule
    
    \multirow{7}{*}{T5-Base$\rightarrow$T5-Base} & FT (I$\rightarrow$O) & \cellcolor{lred} 62.02~($\pm$0.48) & 68.20~($\pm$0.52) \\
    
    & FT (I$\rightarrow$OR) & 56.09~($\pm$0.47) & 57.19~($\pm$0.58) \\
    
    & FT (I$\rightarrow$RO) & 54.60~($\pm$0.66) & 56.76~($\pm$2.74) \\

    & FT (IR$\rightarrow$O) & 41.02~($\pm$1.57) & 66.41~($\pm$0.90) \\
    
    & \methodsp (In) & 55.12~($\pm$2.19) & \cellcolor{lred} 68.45~($\pm$0.83)\\
    
    & \methodsp (Out) & \cellcolor{lblue} 57.26~($\pm$2.68) & \cellcolor{lred} 68.45~($\pm$0.52)\\
    
    & \methodsp (In+Out) & 56.12~($\pm$1.91) & \cellcolor{lblue} 68.41~($\pm$0.99) \\
    
    \bottomrule 
\end{tabular}
}
% \vspace{0.3cm}
\caption{\small \textbf{Failure Analysis}
}
\label{tab:exp:negative}
\vspace{-0.3cm}
\end{wraptable}

Since \methodsp distills FTR knowledge to the student LM, the student's performance is expected to depend on the amount and quality of reasoning knowledge stored in the FTRs.
Thus, to investigate these negative results, we conduct a case study to qualitatively analyze the gold FTRs in OBQA, StrategyQA, ECQA, and QuaRTz.
Overall, we find that FTRs in OBQA and StrategyQA are much more informative than those in ECQA and QuaRTz.
For OBQA and StrategyQA, we find that their gold FTRs tend to have the following properties.
First, they describe a logically sufficient reasoning process for getting from the question (input) to the answer (output).
Second, they provide general and self-contained knowledge that goes beyond the information given in the question and answer.
Meanwhile, FTRs from ECQA and QuaRTz tend to exhibit opposite properties. They usually simply rephrase the question and answer. To illustrate our case study, we give some examples in~\textsection\ref{sec:app:case_study}. To further illustrate how common is simple rephrasing among bad FTRs, we conduct an experiment to quantitatively show it in~\textsection\ref{sec:app:rephrasing}.

Consequently,~\method's failure on ECQA and QuaRTz is owing to the uninformative nature of their human-annotated FTRs.
Given that, we hope future work could annotate datasets with FTRs informative enough, which would collectively convey useful and sufficient reasoning knowledge and thus contribute to performance improvement.

% Building upon \textsection \ref{sec:exp:abl:ftr-quality}, we evaluate various methods on two additional QA datasets: 
% Table~\ref{tab:negative}\zhiyuan{may move it to the Appendix} shows that \methodsp cannot achieve improvements over baselines on them.
% To study the cause, we do a case study on these two datasets' FTRs.
% Some representative examples are shown in Table~\zhiyuan{Add in the Appendix}.
% In many cases, the FTR of ECQA and QuaRTZ simply combines the question and the answer into a statement without providing useful logic that can enhance reasoning.
% Consequently, we argue that \zhiyuan{Not sure what the conclusion here is.}.

% \zhiyuan{Table 2: representative examples. Can Aaron take care of that?}

%\input{sections/8_conclusion}

% \newpage
% \input{sections/6_limitations}
% \input{sections/9_ethicsstatement}

\bibliography{references}
\bibliographystyle{acl_natbib}

\appendix
\section{Appendix}
\label{sec:appendix}

\begin{figure*}
    \centering
    \vspace{-0.9cm}
    \includegraphics[width=\textwidth]{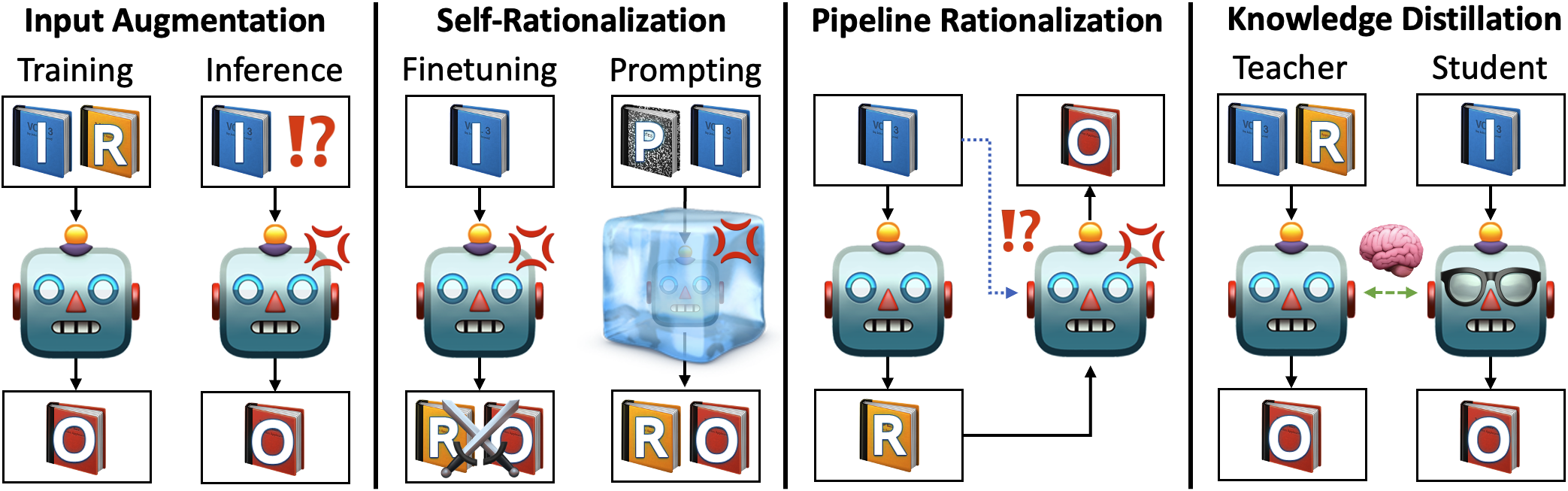}
    % \caption{\textbf{Explanation Tuning Paradigms.} 
    \caption{\textbf{Explanation Tuning Paradigms}
    }
    \vspace{-0.3cm}
    \label{fig:ftr-paradigms}
\end{figure*}

\begin{table}[t]
% \vspace{-0.2cm}
\centering
\scalebox{0.75}{
\begin{tabular}{ccccc}
    \toprule
    \multirow{2}{*}{\textbf{Architecture}} & \multirow{2}{*}{\textbf{Method}} & \multicolumn{2}{c}{\textbf{Accuracy} ($\uparrow$)} \\
    \cmidrule(lr){3-4}
    & & OBQA & StrategyQA \\
    
    \midrule
    
    \multirow{6}{*}{T5-Base$\rightarrow$T5-Base} & FT (I$\rightarrow$O) & 57.93~($\pm$1.15) & 59.05~($\pm$0.23) \\ % & 67.05~($\pm$0.78) \\
    
    & FT (I$\rightarrow$OR) & 50.60~($\pm$1.25) & 53.04~($\pm$1.36) \\
    
    & FT (I$\rightarrow$RO) & 49.93~($\pm$3.20) & 56.18~($\pm$2.58) \\

    & FT (IR$\rightarrow$O) & 48.40~($\pm$1.71) & 49.37~($\pm$2.52) \\

    & FT Dropout (IR$\rightarrow$O) &  58.53~($\pm$1.14) & \cellcolor{lblue} 60.39~($\pm$1.17) \\

    & FT Teacher Init. & \cellcolor{lblue} 59.80~($\pm$1.64) & 59.25~($\pm$0.42) \\
    
    % & \methodsp (In) & 61.07~($\pm$0.12) & \cellcolor{lred} 61.92~($\pm$1.74) \\
    
    % & \methodsp (Out) & \cellcolor{lred} 61.60~($\pm$0.53) & \cellcolor{lblue} 60.72~($\pm$0.20) \\
    
    & \methodsp (In+Out) & \cellcolor{lred} 61.53~($\pm$0.76) & \cellcolor{lred} 61.92~($\pm$1.04) \\

    % \cmidrule(lr){2-5}

    % & CoT Prompting (GPT-3) & I$\rightarrow$RO & X.XX~($\pm$X.XX) & X.XX~($\pm$X.XX) & X.XX~($\pm$X.XX) \\
    
    % & \methodsp Teacher Finetuning & IR$\rightarrow$O & 73.80~($\pm$0.60) & 66.20~($\pm$1.10) \\ %  & 75.85~($\pm$2.20) \\

    \midrule
    
    \multirow{6}{*}{T5-Large$\rightarrow$T5-Large} & FT (I$\rightarrow$O) & 65.60~($\pm$0.40) & 57.58~($\pm$0.70) \\ %  & 74.11~($\pm$1.93) \\

    & FT (I$\rightarrow$OR) & 59.20~($\pm$1.56) & 59.79~($\pm$2.20) \\
    
    & FT (I$\rightarrow$RO) & 59.40~($\pm$0.72) & 55.58~($\pm$1.10) \\

    & FT (IR$\rightarrow$O) & 53.13~($\pm$1.94) & 49.70~($\pm$3.03) \\

    & FT Dropout (IR$\rightarrow$O) & \cellcolor{lblue} 66.87~($\pm$0.31) &  59.85~($\pm$1.80) \\

    & FT Teacher Init. & 66.87~($\pm$1.10) & \cellcolor{lblue} 59.99~($\pm$1.01) \\
    
    % & \methodsp (In) & 67.20~($\pm$0.40) & \cellcolor{lblue} 62.32~($\pm$1.84) \\
    
    % & \methodsp (Out) & \cellcolor{lblue} 68.53~($\pm$1.89) & 62.26~($\pm$0.64) \\
    
    & \methodsp (In+Out) & \cellcolor{lred} 68.73~($\pm$1.55) & \cellcolor{lred} 63.99~($\pm$0.81) \\

    \midrule
    
    % & FT (I$\rightarrow$O) & I$\rightarrow$O & X.XX~($\pm$X.XX) & X.XX~($\pm$X.XX) \\
    
    % & FT (IR$\rightarrow$O) & IR$\rightarrow$O & 53.73~($\pm$2.31) & 49.97~($\pm$2.92) \\ %  & 66.41~($\pm$0.90) \\

    % \multirow{10}{*}{T5-Base} & FT Dropout (IR$\rightarrow$O) & IR$\rightarrow$O & 58.27~($\pm$1.33) & 59.39~($\pm$1.33) \\ %  & 66.41~($\pm$0.90) \\
    
    % & FT & I$\rightarrow$OR & 53.93~($\pm$1.33) & 51.84~($\pm$1.45) \\ %  & 57.19~($\pm$0.58) \\
    
    % & FT & I$\rightarrow$RO & 39.40~($\pm$1.20) & 0.87~($\pm$0.50) \\ %  & 29.38~($\pm$1.85) \\

    \multirow{2}{*}{T5-Large$\rightarrow$T5-Base} & Best T5-Base$\rightarrow$T5-Base & \cellcolor{lblue} 59.80~($\pm$1.64) & \cellcolor{lblue} 60.39~($\pm$1.17) \\

    % \multirow{6}{*}{$\downarrow$} & Non-FTR KD (In) \zhiyuan{Explain} & 31.87~($\pm$2.10) & 53.77~($\pm$0.46) \\

    % \multirow{6}{*}{T5-Base} & Non-FTR KD (Out) & 55.60~($\pm$2.99) & 59.99~($\pm$3.53) \\ %  & 65.43~($\pm$1.22) \\
    
    % & Non-FTR KD (In+Out) & 58.27~($\pm$1.01) & 60.12~($\pm$1.40) \\ %  & \textbf{66.75}~($\pm$0.19) \\

    % & Non-FTR KD (Logit) & 48.53~($\pm$4.06) & 57.72~($\pm$2.12) \\ %  & \textbf{66.75}~($\pm$0.19) \\

    % & \methodsp (In) \zhiyuan{Explain} & 30.07~($\pm$2.97) & 53.91~($\pm$0.69) \\
    
    % & \methodsp (Out) & 55.60~($\pm$2.03) & \cellcolor{lblue} 60.59~($\pm$0.61) \\
    
    & \methodsp (In+Out) & \cellcolor{lred} 60.47~($\pm$0.81) & \cellcolor{lred} 62.39~($\pm$0.42) \\

    \midrule

    GPT-NeoX & CoT (I$\rightarrow$RO) & 33.80 & 55.31 \\

    % & PINTO & IR$\rightarrow$O & 58.85 & 60.87 \\

    \midrule
    
    GPT-3 (text-davinci-003) & CoT (I$\rightarrow$RO) & 86.40 & 66.53 \\
    
    \bottomrule 
\end{tabular}
}
\vspace{0.3cm}
\caption{\small \textbf{\methodsp Main Results (GPT-NeoX FTRs)}
}
\label{tab:exp:main:gpt}
% \vspace{-0.5cm}
\end{table}

\begin{table}[t]
% \vspace{-0.2cm}
\centering
\scalebox{0.75}{
\begin{tabular}{cc}
    \toprule
    {\textbf{Method}} & {\textbf{OBQA Acc.} ($\uparrow$)} \\
    
    \midrule
    
    FT (I$\rightarrow$O) & 44.87~($\pm$0.23) \\
    
    FT (I$\rightarrow$OR) & 38.07~($\pm$1.14) \\
    
    FT (I$\rightarrow$RO) & 38.47~($\pm$2.72) \\

    FT (IR$\rightarrow$O) & 44.80~($\pm$2.23) \\

    FT Dropout (IR$\rightarrow$O) & 47.00~($\pm$1.00) \\

    FT Teacher Init. & 44.20~($\pm$2.95) \\
    
    \methodsp (In) & 47.20~($\pm$1.40) \\
    
    \methodsp (Out) & \cellcolor{lred} 48.13~($\pm$2.12) \\
    
    \methodsp (In+Out) & \cellcolor{lblue} 47.47~($\pm$1.96) \\
    
    \bottomrule 
\end{tabular}
}
\vspace{0.3cm}
\caption{\small \textbf{\methodsp Low-Resource Learning Results}
}
\label{tab:exp:low}
% \vspace{0.3cm}
\end{table}

\subsection{Hyperparameters}
\label{sec:app:hparams}
We always take AdamW as the optimizer.
We stop training when the model performance on the development set has not improved for five epochs.
The maximum epoch is $10$.

For OBQA with T5-Base, we train the teacher model with learning rate of $1\mathrm{e}{-4}$ and batch size of $64$.
We train the student model (by KD) with batch size of $64$.
For OBQA with T5-Large, we train the teacher model with learning rate of $5\mathrm{e}{-5}$ and batch size of $64$.
We train the student model with batch size of $48$.
For the KD training student model, we always search the learning rate in \{$1\mathrm{e}{-4}$, $2\mathrm{e}{-4}$, $3\mathrm{e}{-4}$, $4\mathrm{e}{-4}$, $5\mathrm{e}{-4}$\}. 

For StrategyQA, we always set the warmup rate as $0.06$.
For T5-Base, we train the teacher model with learning rate of $3\mathrm{e}{-4}$ and batch size of $16$.
For T5-Large, we train the teacher model with learning rate of $5\mathrm{e}{-5}$ and batch size of $16$.
For the KD training student model, the batch size is always $16$, and we always search the learning rate in \{$1\mathrm{e}{-4}$, $2\mathrm{e}{-4}$, $3\mathrm{e}{-4}$, $4\mathrm{e}{-4}$, $5\mathrm{e}{-4}$\}.

FT (I$\rightarrow$O), FT (I$\rightarrow$OR), FT (I$\rightarrow$RO), and FT (IR$\rightarrow$O) use the same hyperparameters as the teacher models in the same settings do, except that we always search the learning rate in \{$1\mathrm{e}{-5}$, $2\mathrm{e}{-5}$, $5\mathrm{e}{-5}$ $1\mathrm{e}{-4}$, $2\mathrm{e}{-4}$, $3\mathrm{e}{-4}$, $4\mathrm{e}{-4}$, $5\mathrm{e}{-4}$\}. FT Dropout (IR$\rightarrow$O) uses the same hyperparameters as FT (IR$\rightarrow$O) does, and we search the dropout rate in \{$0.1$, $0.2$, $0.3$, $0.4$, $0.5$, $0.6$, $0.7$, $0.8$, $0.9$\}.  FT Teacher Init. uses the same hyperparameters as FT (I$\rightarrow$O) does.

\subsection{Ablation Studies (Extended)}

We present the full results of ablation studies in the Appendix.
Table \ref{tab:exp:abl:ftr-kd} shows the full results of ablation studies on FTR usage.
Table \ref{tab:exp:abl:ftr-type} shows the full results of ablation studies on FTR perturbation.
Table \ref{tab:exp:abl:ftr-bneck} shows the full results of ablation studies on teacher bottleneck.
Table \ref{tab:exp:abl:task-loss} shows the full results of ablation studies on student task loss.
The details of ablation studies are in \ref{sec:exp:abl}.
In the last three tables, we always use T5-Base$\rightarrow$T5-Base.

\begin{table}[t]
% \vspace{-0.2cm}
\centering
\scalebox{0.75}{
\begin{tabular}{cccc}
    \toprule
    \multirow{2}{*}{\textbf{Method}} & \multicolumn{2}{c}{\textbf{Accuracy} ($\uparrow$)} \\
    \cmidrule(lr){2-3}
    & OBQA & StrategyQA \\
    
    \midrule

    Non-FTR KD (Logit) & 48.53~($\pm$4.06) & 57.72~($\pm$2.12) \\ %  & \textbf{66.75}~($\pm$0.19) \\   

    \midrule

    Non-FTR KD (In) & \cellcolor{lred} 31.87~($\pm$2.10) & \cellcolor{lblue} 53.77~($\pm$0.46) \\

    \methodsp (In) + Gold & \cellcolor{lblue} 31.13~($\pm$2.87) & \cellcolor{lblue} 53.77~($\pm$0.46) \\ %  & 65.43~($\pm$1.22) \\

    \methodsp (In) + GPT-NeoX & 30.07~($\pm$2.97) & \cellcolor{lred} 53.91~($\pm$0.69) \\

    \midrule

    Non-FTR KD (Out) & \cellcolor{lred}55.60~($\pm$2.99) & 59.99~($\pm$3.53) \\ %  & 65.43~($\pm$1.22) \\

    \methodsp (Out) + Gold & \cellcolor{lred} 55.60~($\pm$2.42) & \cellcolor{lred} 61.12~($\pm$0.53) \\ %  & 65.43~($\pm$1.22) \\

    \methodsp (Out) + GPT-NeoX & \cellcolor{lred}55.60~($\pm$2.03) & \cellcolor{lblue} 60.59~($\pm$0.61) \\

    \midrule
    
    Non-FTR KD (In+Out) & 58.27~($\pm$1.01) & 60.12~($\pm$1.40) \\ %  & \textbf{66.75}~($\pm$0.19) \\

    \methodsp (In+Out) + Gold & \cellcolor{lred} 60.93~($\pm$0.12) & \cellcolor{lblue} 61.12~($\pm$2.03) \\ %  & \textbf{66.75}~($\pm$0.19) \\

    \methodsp (In+Out) + GPT-NeoX & \cellcolor{lblue} 60.47~($\pm$0.81) & \cellcolor{lred} 62.39~($\pm$0.42) \\
    
    \bottomrule 
\end{tabular}
}
\vspace{0.3cm}
\caption{\small \textbf{Ablation Study of FTR Usage}
}
\label{tab:exp:abl:ftr-kd}
% \vspace{-0.5cm}
\end{table}

\begin{table}
% \vspace{-0.2cm}
\centering
\scalebox{0.75}{
\begin{tabular}{cccc}
    \toprule
    \multirow{2}{*}{\textbf{FTR Type}} & \multirow{2}{*}{\textbf{\methodsp Variant}} & \multicolumn{2}{c}{\textbf{Accuracy} ($\uparrow$)} \\
    \cmidrule(lr){3-4}
    & & OBQA & StrategyQA \\
    
    \midrule

    Replace & \methodsp (In) & 57.47~($\pm$0.42) & 56.65~($\pm$1.75) \\

    Shuffle & \methodsp (In) & 57.73~($\pm$1.01) & 56.65~($\pm$2.71) \\

    Gold & \methodsp (In) & \cellcolor{lblue} 60.00~($\pm$0.40) & \cellcolor{lblue} 61.39~($\pm$1.90) \\

    GPT-NeoX & \methodsp (In) & \cellcolor{lred} 61.07~($\pm$0.12) & \cellcolor{lred} 61.92~($\pm$1.74) \\

    \midrule
   
    Replace & \methodsp (Out) & 58.67~($\pm$1.10) & 54.31~($\pm$2.84) \\

    Shuffle & \methodsp (Out) & 58.87~($\pm$1.50) & 57.11~($\pm$1.64) \\

    Gold & \methodsp (Out) & \cellcolor{lred} 62.27~($\pm$1.01) &
    \cellcolor{lblue} 59.05~($\pm$1.22) \\

    GPT-NeoX & \methodsp (Out) & \cellcolor{lblue} 61.60~($\pm$0.53) & \cellcolor{lred} 60.72~($\pm$0.20) \\

    \midrule
    
    Replace & \methodsp (In+Out) & 58.87~($\pm$1.30) & 55.44~($\pm$4.43) \\

    Shuffle & \methodsp (In+Out) & \cellcolor{lblue} 59.07~($\pm$0.31) & 56.91~($\pm$1.59) \\

    Gold & \methodsp (In+Out) & \cellcolor{lred} 61.53~($\pm$0.76) & \cellcolor{lblue} 60.45~($\pm$0.31) \\

    GPT-NeoX & \methodsp (In+Out) & \cellcolor{lred} 61.53~($\pm$0.76) & \cellcolor{lred} 61.92~($\pm$1.04) \\

    \midrule

    Replace & \methodsp Teacher & 57.73~($\pm$0.61) & 55.31~($\pm$3.03) \\

    Shuffle & \methodsp Teacher & 56.40~($\pm$1.20) & 56.05~($\pm$2.93) \\

    Gold & \methodsp Teacher & \cellcolor{lblue} 73.80~($\pm$0.60) & \cellcolor{lred} 66.20~($\pm$1.10) \\

    GPT-NeoX & \methodsp Teacher & \cellcolor{lred} 74.33~($\pm$0.46) & \cellcolor{lblue} 64.93~($\pm$1.40) \\
    
    \bottomrule 
\end{tabular}
}
\vspace{0.3cm}
\caption{\small \textbf{Ablation Study of FTR Quality}
}
\label{tab:exp:abl:ftr-type}
% \vspace{-0.5cm}
\end{table}

\begin{table}
% \vspace{-0.2cm}
\centering
\scalebox{0.75}{
\begin{tabular}{cccc}
    \toprule
    \multirow{2}{*}{\textbf{Bottleneck}} & \multirow{2}{*}{\textbf{\methodsp Variant}} & \multicolumn{2}{c}{\textbf{Accuracy} ($\uparrow$)} \\
    \cmidrule(lr){3-4}
    & & OBQA & StrategyQA \\
    
    \midrule

    No & \methodsp (In) & \cellcolor{lblue} 58.67~($\pm$0.70) & \cellcolor{lblue} 49.77~($\pm$2.91) \\

    Yes & \methodsp (In) & \cellcolor{lred} 60.00~($\pm$0.40) & \cellcolor{lred} 61.39~($\pm$1.90) \\

    \midrule

    No & \methodsp (Out) & \cellcolor{lblue} 62.20~($\pm$0.72) & \cellcolor{lred} 59.92~($\pm$0.87) \\

    Yes & \methodsp (Out) & \cellcolor{lred} 62.27~($\pm$1.01) & \cellcolor{lblue} 59.05~($\pm$1.22) \\

    \midrule

    No & \methodsp (In+Out) & \cellcolor{lblue} 58.47~($\pm$0.83) & \cellcolor{lblue} 56.91~($\pm$2.31) \\

    Yes & \methodsp (In+Out) & \cellcolor{lred} 61.53~($\pm$0.76) & \cellcolor{lred} 60.45~($\pm$0.31) \\

    \midrule

    No & \methodsp Teacher & \cellcolor{lblue} 73.40~($\pm$1.51) & \cellcolor{lred} 67.47~($\pm$0.42) \\    

    Yes & \methodsp Teacher & \cellcolor{lred} 73.80~($\pm$0.60) & \cellcolor{lblue} 66.20~($\pm$1.10) \\
    
    \bottomrule 
\end{tabular}
}
\vspace{0.3cm}
\caption{\small \textbf{Ablation Study of Teacher Bottleneck}
}
\label{tab:exp:abl:ftr-bneck}
% \vspace{-0.5cm}
\end{table}

\begin{table}
% \vspace{-0.2cm}
\centering
\scalebox{0.75}{
\begin{tabular}{cccc}
    \toprule
    \multirow{2}{*}{\textbf{Task Loss}} & \multirow{2}{*}{\textbf{\methodsp Variant}} & \multicolumn{2}{c}{\textbf{Accuracy} ($\uparrow$)} \\
    \cmidrule(lr){3-4}
    & & OBQA & StrategyQA \\
    
    \midrule

    Yes & \methodsp (In) & \cellcolor{lblue} 59.73~($\pm$1.10) & \cellcolor{lblue} 56.31~($\pm$1.00) \\

    No & \methodsp (In) & \cellcolor{lred} 60.00~($\pm$0.40) & \cellcolor{lred} 61.39~($\pm$1.90) \\    

    % \cmidrule(lr){2-6}
    \midrule

    Yes & \methodsp (Out) & \cellcolor{lblue} 58.53~($\pm$1.47) & \cellcolor{lblue} 58.65~($\pm$2.02) \\

    No & \methodsp (Out) & \cellcolor{lred} 62.27~($\pm$1.01) & \cellcolor{lred} 59.05~($\pm$1.22) \\

    % \cmidrule(lr){2-6}
    \midrule

    Yes & \methodsp (In+Out) & \cellcolor{lblue} 60.40~($\pm$1.04) & \cellcolor{lblue} 59.32~($\pm$0.35) \\

    No & \methodsp (In+Out) & \cellcolor{lred} 61.53~($\pm$0.76) & \cellcolor{lred} 60.45~($\pm$0.31) \\
    
    \bottomrule 
\end{tabular}
}
\vspace{0.3cm}
\caption{\small \textbf{Ablation Study of Student Task Loss}
}
\label{tab:exp:abl:task-loss}
% \vspace{-0.5cm}
\end{table}

\subsection{Implementation Details}
For T5-Base, the parameter number of a single backbone model is around 220M. As we have two backbone models for the teacher and student model, the total number is around 440M.
For T5-Large, the parameter number of a single backbone model is around 770M, and the total number is around 1.5B.
GPT-NeoX has 20B parameters.

We use NVIDIA Quadro RTX8000 for all experiments, which take around 700 GPU hours.

We implement the models by HuggingFace Transformers. We also heavily use PyTorch and PyTorch Lightning.

\subsection{Case Study of FTR Quality}
\label{sec:app:case_study}

We give a representative example of a good OBQA FTR: ``\textit{\textbf{Question}: There is most likely going to be fog around: \textbf{Answer Choices}: (A) a marsh, (B) a tundra, (C) the plains (D) a desert. \textbf{Gold FTR}: fog is formed by water vapor condensing in the air.}''
The FTR explains the condition for fog formation.
To answer this question with the FTR, one can identify as the answer a place where the condition (water vapor) is strong.
The FTR describes the reasoning for answering the question.

We give a representative example of a bad ECQA FTR: ``\textit{\textbf{Question}: What might a person see at the scene of a brutal killing? \textbf{Answer Choices}: (A) bloody mess, (B) pleasure, (C) being imprisoned, (D) feeling of guilt, (E) cake.
\textbf{Gold FTR}: Bloody mess is covered or stained with blood. A person might see a bloody mess at the scene of a brutal killing.}''
The first sentence of FTR just describes “bloody mess” from its literal meaning.
The second sentence just fills in the answer to the question and rephrases the result into a declarative sentence.
Overall, the FTR uninformatively states “bloody mess” is correct without explaining why.

We give a representative example of a bad QuaRTz FTR: ``\textit{\textbf{Question}: If a moving object slows down, it will have () kinetic energy. \textbf{Answer Choices}: (A) more, (B) less.
\textbf{Gold FTR}: Anything that is moving has kinetic energy, and the faster it is moving, the more kinetic energy it has.}''
The FTR states the outcome of an object speeding up without explanation or reasoning, and the answer is just the opposite of the outcome as the question asks the outcome of an object slowing down.

\subsection{FTR Rephrasing Study}
\label{sec:app:rephrasing}

Bad FTRs tend to simply rephrase the question and answer in an uninformative way (\textsection\ref{sec:exp:failure} and \textsection\ref{sec:app:case_study}).
Interestingly, for such FTRs, humans often can easily get the correct answer given only the FTR without the question.
Based on this observation, we train a T5-Base to do FT (R$\rightarrow$O), \ie to predict the answer taking only the FTR as the input on OBQA and ECQA.
We also conduct this experiment on e-SNLI~\citep{camburu2018snli}, a dataset that has been proven to have very uninformative FTRs~\citep{chen2022rev}.
We omitted StrategyQA and QuaRTz (where an instance has two opposite labels) in this analysis because their FTRs could be used to explain both the correct answer and the opposite label.
The result is shown in Table~\ref{tab:exp:abl:r-o}.\footnote{The accuracy of Random Guessing is the inverse of the number of labels.}
We find that ECQA and e-SNLI tend to do simple rephrasing much more than OBQA does, which further proves the bad quality of ECQA and e-SNLI.\footnote{Note that this experiment is not a comprehensive quantitative analysis as dataset analysis is not our focus.}

\begin{table}
% \vspace{-0.2cm}
\centering
\scalebox{0.75}{
\begin{tabular}{cccc}
    \toprule
    \multirow{2}{*}{Method} & \multicolumn{2}{c}{\textbf{Accuracy} ($\uparrow$)} \\
    \cmidrule(lr){2-4}
    & OBQA & ECQA & e-SNLI \\
    
    \midrule

    Random Guessing & \cellcolor{lblue} 25.00 & \cellcolor{lblue} 20.00 & \cellcolor{lblue} 33.33\\

     FT (R$\rightarrow$O) & \cellcolor{lred} 58.53~($\pm$1.03) & \cellcolor{lred} 97.69~($\pm$0.32) & \cellcolor{lred} 96.03~($\pm$0.42)\\
    
    \bottomrule 
\end{tabular}
}
\vspace{0.3cm}
\caption{\small \textbf{FTR Rephrasing Study}
}
% \vspace{0.3cm}
\label{tab:exp:abl:r-o}
% \vspace{-0.5cm}
\end{table}

\end{document}